\documentclass[twoside,11pt]{article}

\usepackage{jmlr2e}
\usepackage{graphicx}
\usepackage{float}
\usepackage{subfig}
\usepackage{amsmath}
\usepackage{bbm}
\usepackage{slashbox}
\usepackage{array}
\usepackage{multirow}
\usepackage{xcolor}
\usepackage[font={small}, skip=5pt]{caption}
\usepackage{enumitem}

\jmlrheading{1}{2019}{1-42}{NA}{NA}{satyajit001}{Satyajit Neogi and Justin Dauwels}

\ShortHeadings{FLDCRF for Sequence Modeling}{Neogi and Dauwels}
\firstpageno{1}

\begin{document}

\title{Factored Latent-Dynamic Conditional Random Fields for Single and Multi-label Sequence Modeling}

\author{\name Satyajit Neogi \email satyajit001@e.ntu.edu.sg \\
       \addr School of Electrical and Electronic Engineering\\
       Nanyang Technological University\\
       50 Nanyang Avenue, Singapore - 639798
       \AND
       \name Justin Dauwels \email jdauwels@ntu.edu.sg \\
       \addr School of Electrical and Electronic Engineering\\
       Nanyang Technological University\\
       50 Nanyang Avenue, Singapore - 639798}

\editor{}

\maketitle

\begin{abstract}


Conditional Random Fields (CRF) are frequently applied for labeling and segmenting sequence data. \citet{LDCRF} introduced hidden state variables in a labeled CRF structure in order to model the latent dynamics within class labels, thus improving the labeling performance. Such a model is known as Latent-Dynamic CRF (LDCRF). We present Factored LDCRF (FLDCRF), a structure that allows multiple latent dynamics of the class labels to interact with each other. Including such latent-dynamic interactions leads to improved labeling performance on single-label and multi-label sequence modeling tasks. We apply our FLDCRF models on two single-label (one nested cross-validation) and one multi-label sequence tagging (nested cross-validation) experiments across two different datasets - UCI gesture phase data and UCI opportunity data. FLDCRF outperforms all state-of-the-art sequence models, i.e., CRF, LDCRF, LSTM, LSTM-CRF, Factorial CRF, Coupled CRF and a multi-label LSTM model in all our experiments. In addition, LSTM based models display inconsistent performance across validation and test data, and pose difficulty to select models on validation data during our experiments. FLDCRF offers easier model selection, consistency across validation and test performance and lucid model intuition. FLDCRF is also much faster to train compared to LSTM, even without a GPU. FLDCRF outshines the best LSTM model by $\sim$4\% on a single-label task on UCI gesture phase data and outperforms LSTM performance by $\sim$2\% on average across nested cross-validation test sets on the multi-label sequence tagging experiment on UCI opportunity data. The idea of FLDCRF can be extended to joint (multi-agent interactions) and heterogeneous (discrete and continuous) state space models.

\end{abstract}


\section{Introduction}  \label{sec:intro}

Labeling and segmenting sequence data is a very common problem of machine learning, which has various applications in the fields of Natural Language Processing (e.g., noun phrase chunking \citep{DCRF, NLP-scratch}, Part of Speech tagging \citep{DCRF, NLP-scratch}, named entity recognition \citep{NER-neural, NLP-scratch} etc.), computer vision (gesture recognition \citep{LDCRF}, activity recognition \citep{Opportunity-action-recognition} etc.), and information extraction \citep{MEMM}. Given an observed input sequence $\textbf{x} = \{x_t\}_{t=1:T}$, the labeling task aims at automatically assigning labels $y_t$ for each input value $x_t$.

\begin{figure}[h]%
\centering
\subfloat[\scriptsize{LCCRF}]{{\includegraphics[scale=0.5]{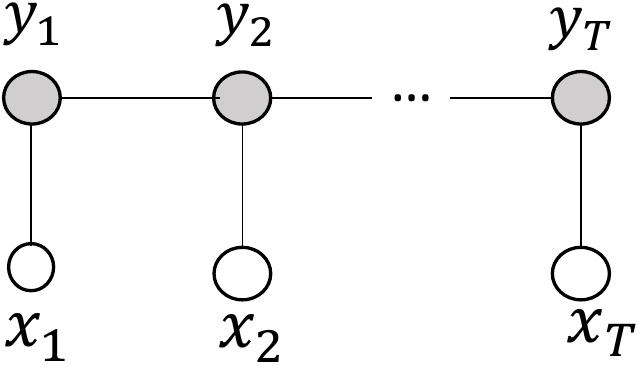} }}%
\qquad
\subfloat[\scriptsize{HCRF}]{{\includegraphics[scale=0.5]{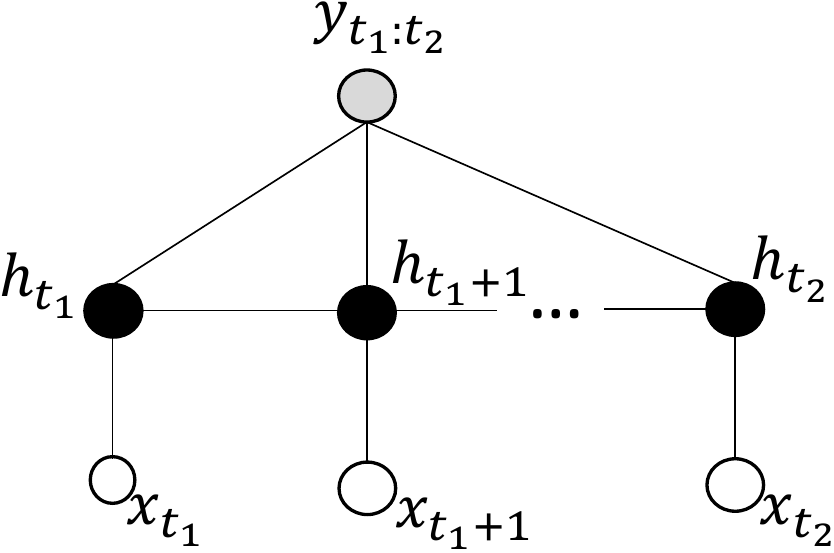} }}%
\qquad
\subfloat[\scriptsize{LDCRF}]{{\includegraphics[scale=0.5]{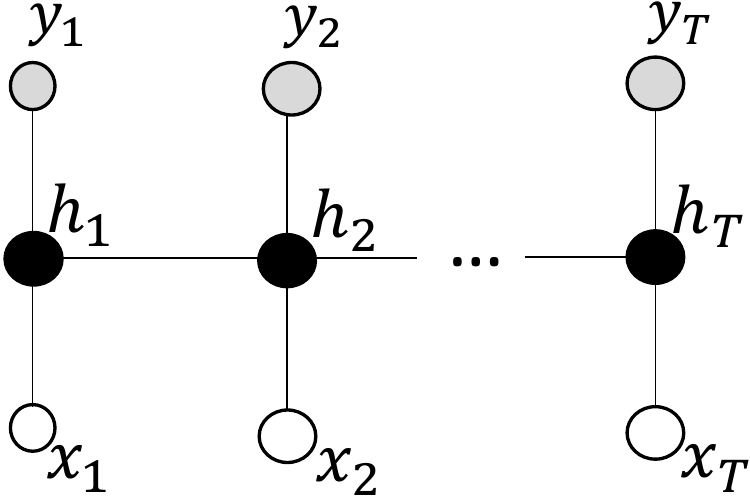} }}%
\caption{Single-label sequential CRF variants. White nodes are (training + testing) observed and black nodes are hidden. Grey nodes are observed only during training.}%
\label{fig:CDCRF}%
\end{figure}

Conditional Random Fields (CRF, \citet{CRF}) are frequently applied to the sequence labeling task. The simplest of its kind is the Linear Chain Conditional Random Field (LCCRF, also called CRF), which is defined over $\{x_t\}_{t=1:T}$ and $\{y_t\}_{t=1:T}$ by the graphical constraints depicted in Fig. \ref{fig:CDCRF}a. The output sequence values ($y_t$) usually belong to a predefined set $\mathcal{Y}$ of class labels. For example, in a continuous human  activity recognition problem, $\mathcal{Y}$ = $\{\text{`sitting', `standing', `walking', `lying'}\}$. Due to the Markov dependency assumption among the label sequence $\{y_t\}_{t=1:T}$, a LCCRF models the extrinsic dynamics (temporal dependency) within the class labels. 

In addition to the extrinsic dynamics among the class labels, each class in $\mathcal{Y}$ has underlying intrinsic dynamics. For instance, each class `sitting', `standing', `walking' etc. contains temporal transitions between certain hidden posture states. \citet{LDCRF} introduced a layer of hidden variables $\{h_t\}_{t=1:T}$ (see Fig. \ref{fig:CDCRF}c) in a LCCRF structure, in order to model a complete latent (both extrinsic and intrinsic) dynamics of the class labels. Each class in $\mathcal{Y}$ is associated with distinct sets of hidden states. Transitions among the states (to capture the intrinsic and extrinsic dynamics) are made possible through the layer of hidden variables $\{h_t\}_{t=1:T}$. LDCRF has been shown to outperform popular sequence models viz., Hidden Markov Models, LCCRF and Hidden-state CRFs (see Section \ref{sec:literature}) on several sequence labeling tasks \citep{LDCRF, LDCRF-CRF}. We describe the LDCRF model in Appendix A.

\begin{figure}[h]%
\centering
\subfloat[\scriptsize{HMM}]{{\includegraphics[scale=0.6]{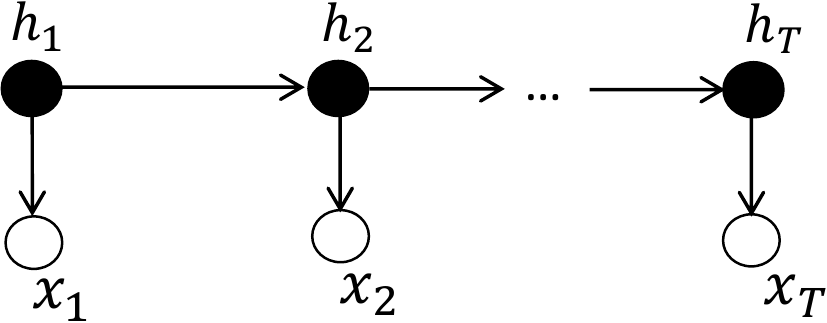} }}%
\hspace{4em}
\subfloat[\scriptsize{Factorial HMM}]{{\includegraphics[scale=0.6]{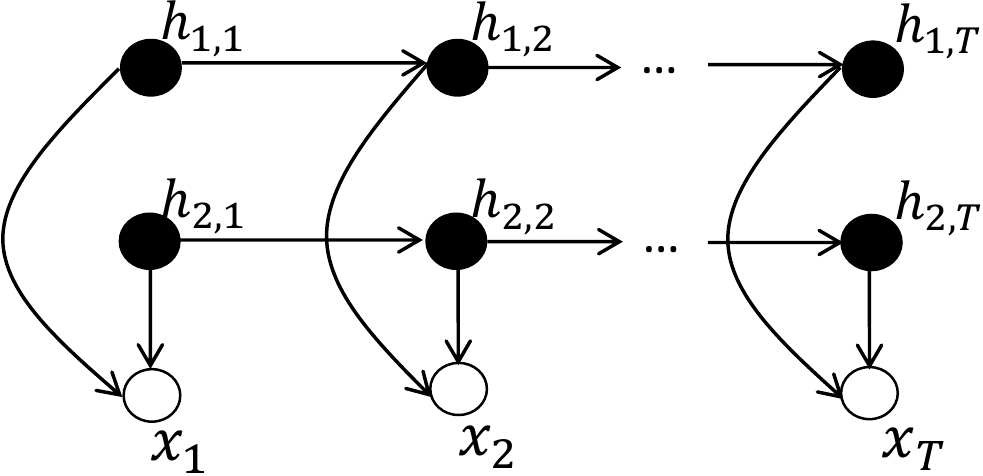} }}%
\caption{(a) A Hidden Markov Model. (b) A Factorial HMM with multiple cotemporal state variables.}%
\label{fig:HMM-FHMM}%
\end{figure}

Learning performance in a LDCRF can only be improved by varying the number $N_s$ of states associated to each class label. This only mode of variation restricts the model capabilities, and results in a rapidly growing state transition matrix with size $(N_l \cdot N_s) \times (N_l \cdot N_s)$, where $N_l = |\mathcal{Y}|$. Such increment results in greater model complexity and requires more training data. A solution for a Hidden Markov Model with a similar structure of the state space was proposed by \citet{Factorial-HMM}, where multiple cotemporal state variables ($h_{1,t}$, $h_{2,t}$ etc.) replaced a single hidden state variable (see Fig. \ref{fig:HMM-FHMM}). The hidden states are distributed among the state variables and all state variables assumed independent temporal dynamics over the input observations \{$x_t\}_{t=1:T}$, resulting in reduction in number of state transition parameters.

\begin{figure}[h]%
\centering
\subfloat[\scriptsize{LDCRF}]{{\includegraphics[scale=0.6]{LDCRF.pdf} }}%
\hspace{4em}
\subfloat[\scriptsize{FLDCRF-s}]{{\includegraphics[scale=0.6]{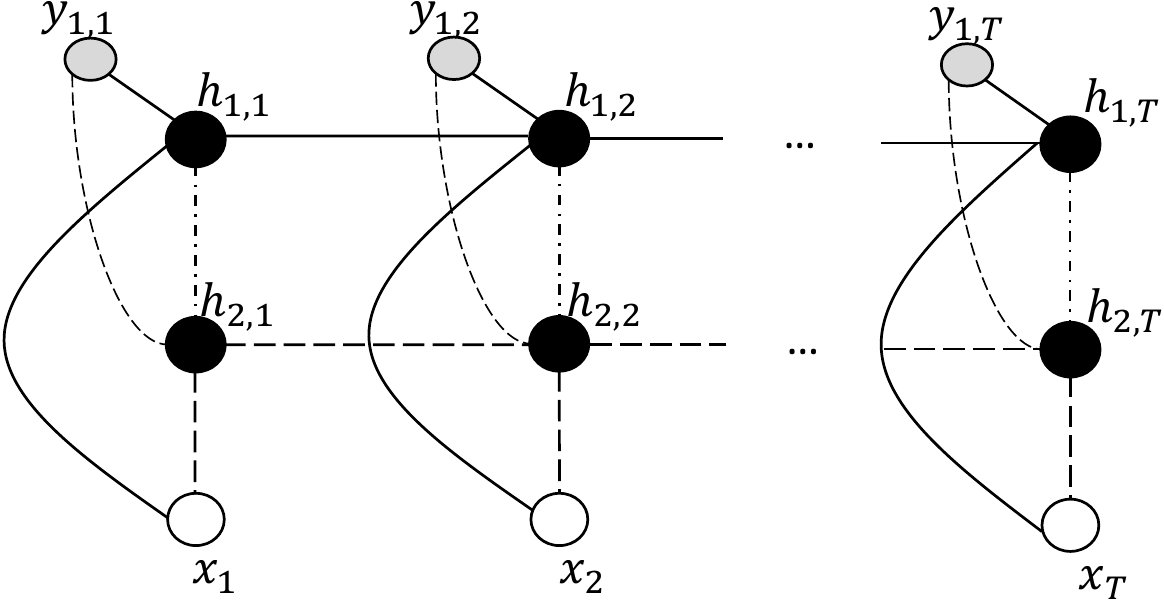} }}%
\caption{(a) A Latent-Dynamic Conditional Random Field. (b) FLDCRF graphical model for single-label sequence prediction. Solid and dashed connections depict two different latent dynamics within class labels \{$y_t\}_{t=1:T}$. Dashed-dotted inter-layer connections capture interactions among different latent dynamics. The graph shows only two hidden layers and Markov connections for transitions between state variables \{$h_{i,t}\}_{t=1:T}$ in layer $i$, $i = 1:L$. More hidden layers and long-range dependencies (semi-Markov for transitions and Markov/semi-Markov for dashed-dotted inter-layer influences) are also possible but omitted for simplicity.}%
\label{fig:LDCRF-FLDCRFs}%
\end{figure}

In addition, it is possible that there exists multiple interacting latent dynamics within the class labels. For example, in a continuous human activity recognition problem (e.g., with four classes, `lying', `sitting', `standing' and `walking'), a human being has two interacting dynamics - one along the plane and another in vertical direction. Such interaction among different latent dynamics can be captured by connecting state variables ($h_{1,t}$, $h_{2,t}$ etc.) across layers (see Fig. \ref{fig:LDCRF-FLDCRFs}b). LDCRF (see Fig. \ref{fig:LDCRF-FLDCRFs}a) ignores any possible interaction among the associated hidden states at slice~$t$, and thus is unable to capture such interactions. Considering all these limitations of LDCRF for a single label sequence modeling and inspired by Factorial HMM, we present a generalization of LDCRF, called Factored Latent-Dynamic Conditional Random Fields (FLDCRF, \citet{mypaper}), in order to:

\begin{itemize}
\item{generate new models by varying the number of hidden layers,}
\item{generate factorized models \citep{Factorial-HMM} with fewer parameters, which need less data for training, and}
\item{model mutliple interacting latent dynamics within class labels.}
\end{itemize}

We denote the FLDCRF single-label sequence model (see Fig. \ref{fig:LDCRF-FLDCRFs}b) as FLDCRF-s. FLDCRF-s improves LDCRF performance across four nested cross-validation experiments on UCI gesture phase data \citep{gesture-phase-data} and UCI opportunity data \citep{UCI-oppor} (see Section \ref{sec:Results}). 

LSTMs \citep{LSTM} are the most popular kind of Recurrent Neural Networks (RNN) for sequence modeling, largely due to their ability to capture long-range dependencies among sequence values. They are frequently applied to sequence labeling tasks, viz., named entity recognition \citep{NER-neural, BILSTM-CRF}, noun phrase chunking \citep{BILSTM-CRF}, action recognition \citep{action-lstm} etc. A variant of LSTM, combined with a CRF layer \citep{NER-neural} for classification, has recently been very successful in sequence labeling tasks \citep{NER-neural, BILSTM-CRF, LSTM-Med-labeling}. We compare FLDCRF-s with both LSTM and LSTM-CRF on the test data in all our experiments. In addition, we analyze several important modeling aspects of FLDCRF-s and LSTM, viz., ease of model selection, consistency across validation and test performance, computation times for training and inference etc. FLDCRF-s not only outperforms LSTM and LSTM-CRF on test data across all experiments, but also offers easier model selection and more consistent performance across validation and test data (see Section \ref{sec:Results}). In addition, FLDCRF-s requires notably less training and inference times than LSTM models, even without GPU implementation.

Multi-task sequence learning \citep{DCRF, NLP-scratch} is the task to jointly tag sequence values with multiple label categories. Such learning is often helpful where different label categories are related to each other. For example, let us consider the two tasks of continuously recognizing a high-level action (e.g., with two classes, `relaxing' and `exercising') and a low-level action (e.g., with four classes, `lying', `sitting', `standing' and `walking') over some given input (sensor) sequence data. Very clearly the two action types are related and a joint modeling of both is likely to improve the individual recognition performances with the help of additional contextual information (see Section \ref{sec:Results}).

\begin{figure}[h]%
\centering
\subfloat[\scriptsize{Coupled CRF \citep{multilabel-sequence-3}.}]{{\includegraphics[scale=0.5]{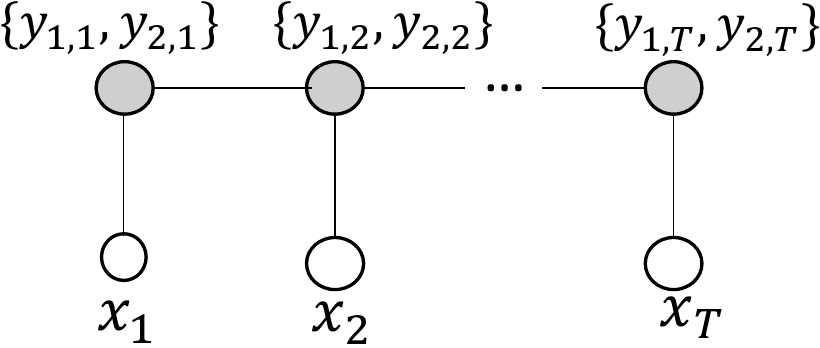} }}%
\hspace{2em}
\subfloat[\scriptsize{Factorial CRF \citep{DCRF}.}]{{\includegraphics[scale=0.5]{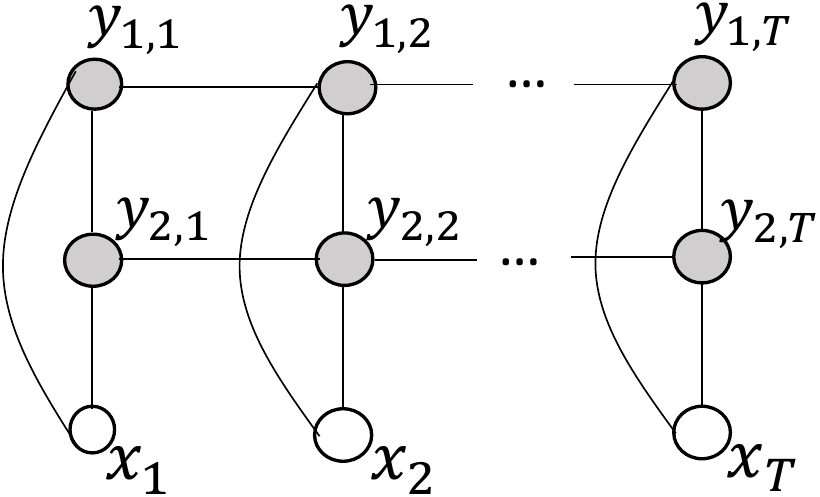} }}%
\hspace{2em}
\subfloat[\scriptsize{Coupled CRF factored graphical representation.}]{{\includegraphics[scale=0.5]{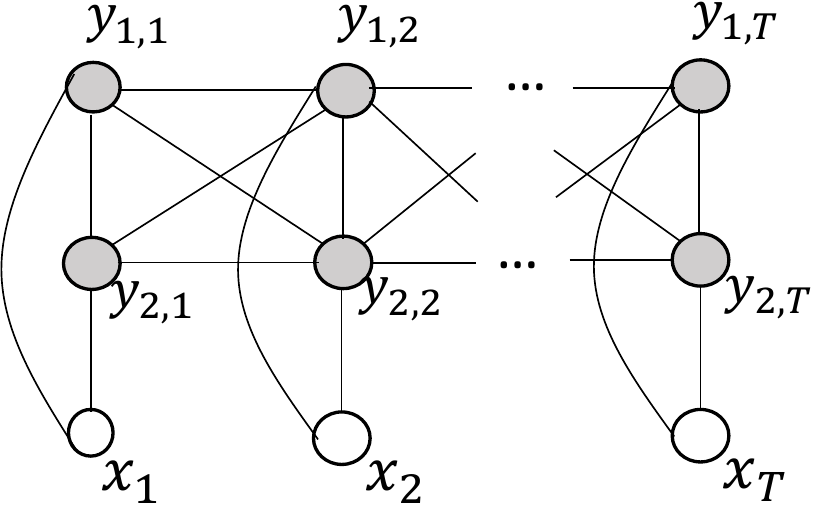} }}%
\caption{Existing multi-label sequence learning models. (b) and (c) are examples of Dynamic Conditional Random Fields \citep{DCRF}.}%
\label{fig:multi-label-existing}%
\end{figure}

Dynamic CRFs (see Fig. \ref{fig:multi-label-existing}b-c, \citet{DCRF}) are very popular \citep{DCRF, DCRF-appl1} for joint modeling of multiple label sequences. Different label categories $y_{1,t}$, $y_{2,t}$ etc. interact with each other through inter-layer links. The structure in Fig. \ref{fig:multi-label-existing}b is popularly known as a Factorial CRF.

\begin{figure}[h]%
\centering
\subfloat[\scriptsize{FLDCRF-m}]{{\includegraphics[scale=0.55]{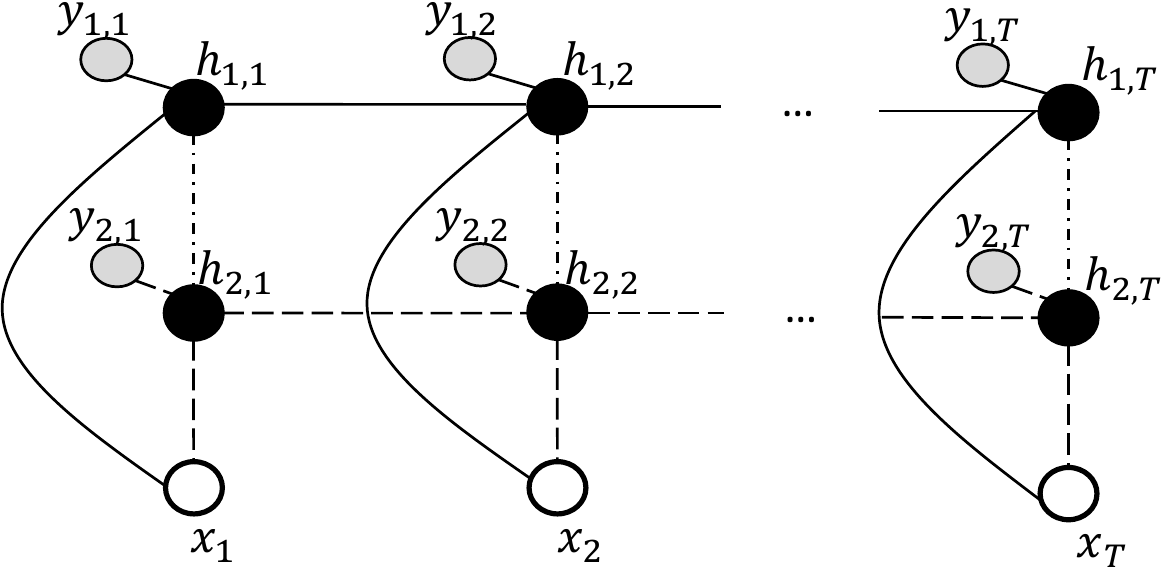} }}%
\qquad
\subfloat[\scriptsize{FLDCRF-i}]{{\includegraphics[scale=0.55]{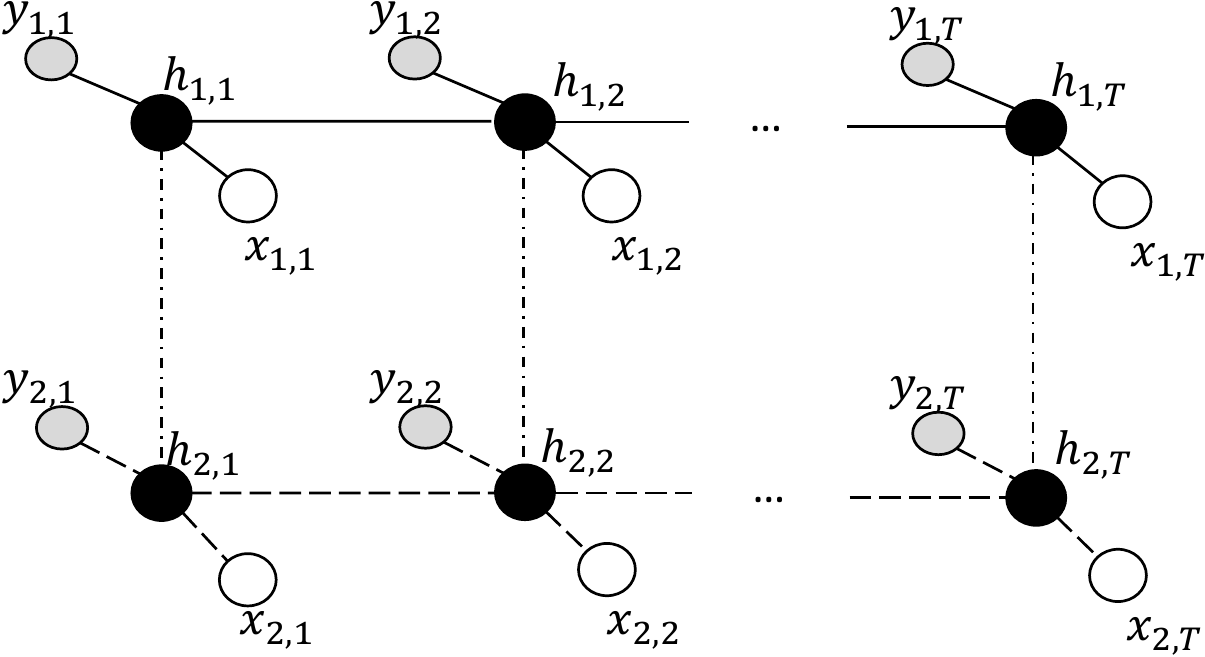} }}%
\caption{Variants of FLDCRF. (a) FLDCRF graphical model for multi-label sequence prediction. Different label categories, $y_{1,t}$ and $y_{2,t}$, over input $x_t$ are connected through their respective hidden layers, $h_{1,t}$ and $h_{2,t}$, influencing each others' latent dynamics. (b) FLDCRF graphical model for social interaction. Two different objects, ($\{x_{1,t}\}$, $\{y_{1,t}\}$) and ($\{x_{2,t}\}$, $\{y_{2,t}\}$) are shown to be interacting in the latent space through their hidden layers to effect each others' dynamics. Longer range dependencies are possible in both models but are omitted for simiplicity. In fact, we add markov connections for influence links in our multi-label experiment (see Section \ref{sec:Experiments}).}%
\label{fig:FLDCRF-mi}%
\end{figure}

Similar to a Linear chain CRF, a DCRF does not contain any hidden state variables and therefore is unable to capture the latent (intrinsic and extrinsic) dynamics within the classes of different label categories (e.g., high and low-level actions). Moreover, connections between the latent dynamics of different label categories can possibly capture deeper understanding of their interactions. Considering such limitation of DCRF, we propose a multi-label variant of FLDCRF (FLDCRF-m, see Fig. \ref{fig:FLDCRF-mi}a), in order to:

\begin{itemize}
\item{introduce latent variables in a general DCRF structure,}
\item{model interactions among the latent dynamics of two or more inter-related label categories, and}
\item{provide a general representation of DCRFs with embedded hidden state variables for each output variable. The original DCRF structure can be recovered by associating one hidden state per class of the output variables (see Section \ref{subsec:FLDCRF-model} for mathematical details).}
\end{itemize}

A less popular CRF variant for joint sequence tagging is coupled CRF (CCRF, see Fig. \ref{fig:multi-label-existing}, \citet{multilabel-sequence-3}), that utlizes a combined state space for different label categories $y_{1,t}$, $y_{2,t}$ etc. Recurrent Neural Networks \citep{multilabel-sequence-2} have also been applied to joint tagging of sequences. We present a structured and detailed review of literature on joint sequence labeling in Section \ref{sec:literature}.

FLDCRF-m outperforms Factorial CRF and coupled CRFs on our multi-task sequence labeling experiment on the UCI opportunity dataset \citep{UCI-oppor} (see Section \ref{sec:Results}). FLDCRF-m also outshines all tested single-label sequence models (trained and inferred separately on different label categories), viz., CRF, LDCRF, FLDCRF-s, LSTM and LSTM-CRF on both tasks combined, while improving the individual labeling performances on most nested test sets. We also compare FLDCRF-m with a multi-task LSTM model, which we refer to as LSTM-m in the paper, similar to the BiRNN model by \cite{multilabel-sequence-2}.

Another variant of FLDCRF, called the interaction variant (FLDCRF-i, see Fig. \ref{fig:FLDCRF-mi}b), captures relationship among the dynamics of multiple agents in a social environment. FLDCRF-i can be applied to multi-agent sequence modeling and prediction tasks in a social environment, e.g., joint intention prediction of multiple pedestrians). Social LSTM \citep{social-LSTM} provides a similar unsupervised framework for joint path prediction. We present a general mathematical description of all the 3 FLDCRF model variants (FLDCRF-s, FLDCRF-m and FLDCRF-i) in Section \ref{sec:Models}.

We summarize the main contrtibutions of this paper below:

\begin{itemize}
\item We propose a generalization of LDCRF, called Factored LDCRF, to capture sequential interactions in latent space.
\item Single label variant of FLDCRF (FLDCRF-s) allows multiple latent dynamics of the class labels to interact with each other, leading to deeper understanding. FLDCRF-s outperforms CRF, LDCRF, LSTM and LSTM-CRF models in the single-label sequence tagging experiments.
\item Multi-label FLDCRF (FLDCRF-m) outperforms Factorial CRF, coupled CRF, multi-label LSTM models and all considered single label sequence models (CRF, LDCRF, LSTM and LSTM-CRF) on the joint sequence tagging task. We show that while factorial CRF and coupled CRF do not yield much improvement over a CRF, the interaction among the latent dynamics in FLDCRF-m results in substantial improvements (upto 3\%) over the individual LDCRF models.
\item In addition to model performance on test data, we compare FLDCRF and LSTM models on several important modeling aspects, such as, ease of model selection on the validation data, consistency across validation and test performance, and computation times for training and inference (see Section \ref{sec:Results}).
\end{itemize}

The rest of the paper is organized as follows. In Section \ref{sec:literature}, we review existing literature on single and multi-label sequence modeling. In Section \ref{sec:Models}, we describe the proposed Factored Latent-Dynamic Conditional Random Fields (FLDCRF) and explain its training and inference mechanisms. In Section \ref{sec:Dataset}, we describe the datasets considered in our experiments. We discuss our experimental setup (feature preprocessing, models tested, metrics etc.) in Section \ref{sec:Experiments} and present our results in Section \ref{sec:Results}. We briefly summarize several modeling aspects of FLDCRF and LSTM in Section \ref{sec:Discussion}. Finally, we offer concluding remarks and ideas for future research in Section \ref{sec:conclude}.

\section{Literature Review} \label{sec:literature}

We discuss existing literature on single and multi-label sequence modeling in this Section.

Approaches to sequence labeling can be broadly classified into two categories: a) Generative (e.g., Hidden Markov Models \citep{HMM}), which learn the joint distribution $P(\textbf{y}, \textbf{x})$ from the training data; and b) Discriminative (e.g., Conditional Random Fields \citep{CRF}, Maximum Entropy Markov Models \citep{MEMM} etc.), which directly learn the conditional distribution $P(\textbf{y} \mid \textbf{x})$ from the training data. Discriminative models have often been shown to outperform generative models on sequence labeling tasks \citep{CRF, LDCRF, MEMM}. In particular, Conditional Random Fields (CRF) are frequently applied to the sequence labeling problem \citep{BILSTM-CRF, NLP-scratch, NER-neural}.

A simple CRF model, also called Linear Chain CRF (LCCRF, see Fig. \ref{fig:CDCRF}a), is defined directly over the input (\{$x_t\}_{t=1:T}$) and output (\{$y_t\}_{t=1:T}$) sequence variables. The class labels within a sequence (\{$y_t\}_{t=1:T}$) have an extrinsic dynamics (temporal dependency), as well as there exists underlying intrinsic dynamics within each class. For example, in a continuous activity recognition problem with classes `sitting', `standing', `walking' etc., each class contains temporal transitions between certain posture states. Since such posture states are hidden, the intrinsic dynamics within a particular class can be captured by associating hidden (latent) states and allowing their transitions. The simple CRF model captures the extrinsic dynamics among the class labels through its transition links (connecting $y_{t-1}$ and $y_t$). However, it does not have any latent variables in its structure and thus is unable to capture the intrinsic dynamics within the class labels.

To address this issue, Hidden-state Conditional Random Fields (HCRF, \citet{HCRF}) introduce hidden variables in a labeled CRF structure (see Fig. \ref{fig:CDCRF}b), in order to capture the intrinsic dynamics within the class labels. The hidden variables (\{$h_t\}_{t=t_1:t_2}$) can assume values from a predefined set of hidden states (usually specified by numbers) and the Markov dependency among the variables allows to capture the intrinsic dynamics within the class label $y_{t_1:t_2}$. However, such a structure requires prior segmentation of the training sequences according to the class labels and models each class label separately (similar to unsupervised HMMs for each class label, only trained by a discriminative approach). Hence, HCRF fails to capture the extrinsic dynamics between the class labels. 

An alternative approach, named Latent-Dynamic Conditional Random Fields (LDCRF), was proposed by \citet{LDCRF} (see Fig. \ref{fig:CDCRF}c). In a LDCRF, prior segmentation of the training sequences is not necessary. Thus, LDCRF is able to capture both intrinsic and extrinsic dynamics of the class labels. LDCRF restricts the states associated to each class label to be disjoint. Each hidden state variable $h_t$ is also constrained to belong to the states associated to the class label $y_t$. These two constraints help to keep the model computations (during training and inference) tractable, despite the insertion of hidden variables. LDCRF has been shown to outperform HMM, CRF and HCRF on several sequence labeling tasks \citep{LDCRF, LDCRF-CRF}. \citet{embedded-CRF} recently proposed a very similar model called Embedded-State Latent CRFs. They claim to factorize the log potential as the novelty over a LDCRF, however such factorization is not reflected in their model structure and mathematical descriptions.

We propose FLDCRF-s in order to model multiple interacting latent dynamics within the class labels (see Section \ref{sec:intro}). We describe how FLDCRF-s mathematically generalizes a LDCRF in Section \ref{sec:Models}. We demonstrate improvement in performance by FLDCRF-s over LDCRF across nested cross-validation experiments on two different datasets (see Section \ref{sec:Results}). 

LSTMs \citep{LSTM} are the most popular kind of Recurrent Neural Networks (RNN) for sequence modeling. They are frequently applied to sequence labeling tasks, e.g., named entity recognition \citep{NER-neural, BILSTM-CRF}, noun phrase chunking \citep{BILSTM-CRF}, action recognition \citep{action-lstm} etc. LSTM-CRF, a variant of LSTM, combined with a CRF layer \citep{NER-neural} for classification, has recently been very popular for sequence labeling \citep{NER-neural, BILSTM-CRF, LSTM-Med-labeling}. 

A few studies \citep{LDCRF-LSTM-1, LDCRF-LSTM-2, LDCRF-LSTM-3} have compared LDCRF and LSTM on the same classification tasks. While \citet{LDCRF-LSTM-2, LDCRF-LSTM-3} show LSTM and its variants to outperform LDCRF, \citet{LDCRF-LSTM-1} present results with each outperforming the other on different tasks. We show that although LSTM and LSTM-CRF outperform LDCRF on certain experiments, FLDCRF-s outperforms both LSTM and LSTM-CRF across all our experiments. By contrast, the best LSTM (and LSTM-CRF) models outshine the best FLDCRF-s performance on most of the validation sets across experiments in the paper. However, the same LSTM models (with optimized hyperparameters), kept unchanged or retrained on the validation+train data, fail to beat the optimized FLDCRF-s models on the test data across experiments (see Section \ref{sec:Results}). In addition to this inconsistent validation and test performance, LSTM models exhibit large variations in validation performance across different hyperparameter settings, with quite a few models producing significantly lesser validation performance. This variation exists without any easily discernible patterns among different hyperparameter settings. The variable performance across validation sets, lower worst cases, and inconsistency in validation and test performance make it quite tedious to select appropriate LSTM models, and raise concerns about the stability of the models in practical deployment. On the other hand, FLDCRF models with its intuitive structure, brings consistency in validation and test performance and offers a much easier model selection process. LSTM models also take significantly longer to train than FLDCRF-s models.  We present results to support all the above statements in Section \ref{sec:Results}.

\begin{figure}[h]%
\centering
\subfloat[\scriptsize{Coupled LDCRF}]{{\includegraphics[scale=0.5]{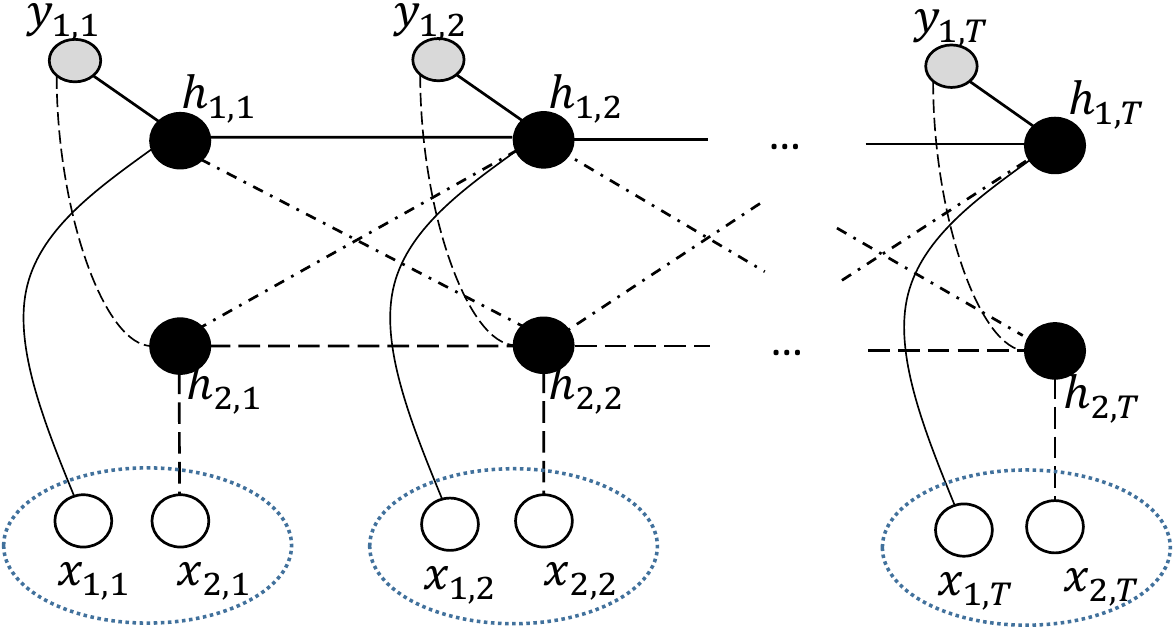} }}%
\hspace{4em}
\subfloat[\scriptsize{Linked LDCRF}]{{\includegraphics[scale=0.5]{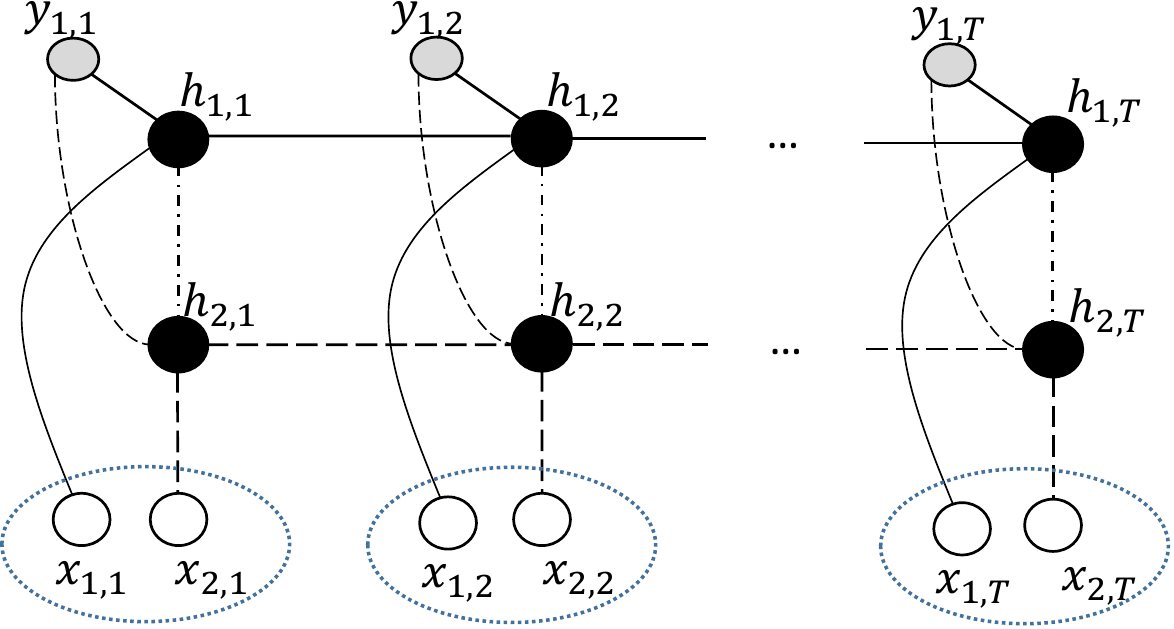} }}%
\caption{Multi-view LDCRFs \citep{Coupled-LDCRF}. The features $x_t$ in Fig. \ref{fig:LDCRF-FLDCRFs}b are distributed across two variables $x_{1,t}$ and $x_{2,t}$, i.e., $x_t = \{x_{1,t}, x_{2,t}\}$.}%
\label{fig:CLDCRF}%
\end{figure}

A factored variant of LDCRF, called multi-view LDCRF (MVLDCRF, see Fig. \ref{fig:CLDCRF}, \citet{Coupled-LDCRF}), was proposed to capture interactions among different latent dynamics arising due to different kinds of input features within same class labels. A MVLDCRF is a special case of FLDCRF-s, where different latent dynamics within class labels $\{y_t\}_{t=1:T}$ are learned from disjoint feature subsets $\{x_{1,t}\}_{t=1:T}$ and \{$x_{2,t}\}_{t=1:T}$ of $x_t = \{x_{1,t}, x_{2,t}\}$. \citet{Coupled-LDCRF} showed a MVLDCRF to outperform a LDCRF trained with the entire feature set $x_t$ in their experiments. However, such explicit distribution of features in order to force the model to learn different latent dynamics does not gurantee improved performance, as we demonstrate by our experiments (see Section \ref{subsec:multi-view}). In some cases, performance of such a MVLDCRF falls in between those of the individual LDCRFs, i.e., one trained with $x_{1,t}$ and another with $x_{2,t}$. These results are not reported by \citet{Coupled-LDCRF}. In most other cases, a LDCRF trained with the entire feature set $x_t$ outperforms MVLDCRF and individual LDCRFs. By contrast, a FLDCRF-s with all hidden layers modeling their dynamics from the entire $x_t$ lets the model to learn different latent dynamics that may exist, and in general improves the LDCRF (with $x_t$) performance. In the case a feature distribution ($x_{1,t}$, $x_{2,t}$ etc.) is available, it is advisable to obtain individual FLDCRF-s performances on the individual feature subsets, as well as on entire $x_t$, and select the best performance. It is better to avoid learning from distributed features on the same training labels as in MVLDCRF, as we demonstrate on 10 different test sets in Section \ref{subsec:multi-view}.

Multi-task learning \citep{Multi-task} is a very common problem of machine learning, where multiple label categories are trained and/or inferred jointly over certain input features. Modeling multiple labels together serves additional context information to individual labeling tasks, which help to improve the overall labeling performance. However, the model must allow appropriate label interactions for beneficial results. Multi-task learning is frequently applied to tag images with multiple labels \citep{Multi-label-image-1, Multi-label-image-2}.

Multi-task sequence learning \citep{DCRF, NLP-scratch} is the task to jointly tag sequence values with multiple label categories. Exisitng multi-task sequence labeling approaches can be broadly classified into two categories: 

\begin{enumerate}
\item{\textbf{Using a joint state variable} (see Fig. \ref{fig:multi-label-existing}a), i.e.,  $y_t$ = $\{y_{1,t}, y_{2,t}\}$} \citep{multilabel-sequence-3, multilabel-sequence-4, multilabel-sequence-2}. \citet{multilabel-sequence-3} learned a coupled CRF (CCRF) model that allowed the two (or more) label categories to interact temporally through their coupled transition matrix. \citet{multilabel-sequence-4} proposed a model called JERL (Joint Entity Recognition and Linking) for joint named entity recognition and disambiguation, which is similar to a coupled CRF. A correlated BiRNN model was introduced by \citet{multilabel-sequence-2}, where multiple label categories (punctuation and capitalization) were combined to pass through the same RNN hidden layer, thus capturing the multi-label interactions through the temporal transitions of the hidden layer. Although all such joint models showed improvements over the individual models, the joint state space makes the model too large with multiple classes (or latent states) for each task. 

\item{\textbf{Using separate state variables} ($y_{1,t}$, $y_{2,t}$ etc.) \textbf{with cotemporal links} (see Fig. \ref{fig:multi-label-existing}b-c \citet{DCRF}) to capture interaction. These structures are derived from Dynamic CRFs proposed by \citet{DCRF}. The structure in Fig. \ref{fig:multi-label-existing}b is also well known as a Factorial CRF (FCRF). These factored representations from a DCRF require fewer parameters than modeling with the joint state variable, and allow easy addition/removal of links depending on required model complexity and available training data. For instance, the coupled CRF model by \citep{multilabel-sequence-3} can be efficiently representled by Fig. \ref{fig:multi-label-existing}c, with substantailly reduced model parameters for the same training labels $\{y_{1,t}\}_{t=1:T}$ and $\{y_{2,t}\}_{t=1:T}$. \citet{DCRF} showed improvements in labeling accuracy with a FCRF while considering a joint noun phrase (NP) chunking and part of speech (POS) tagging task. They also reported better performance for FCRF compared to two separate CRFs for reduced training data.}
\end{enumerate}

Several other studies \citep{NLP-scratch, multilabel-sequence-1, multilabel-sequence-5, multilabel-sequence-6, multilabel-sequence-action} have been conducted on joint sequence labeling. \citet{multilabel-sequence-6} proposed a dual layer CRF for joint decoding of different tasks (segmentation and POS tagging), however showed marginal improvement over individual labeling with their approach. \citet{multilabel-sequence-5} designed a Multi-CRF with a shared entropy based training likelihood function. \citet{NLP-scratch} and \citet{multilabel-sequence-1} reviewed existing approaches to multi-task sequence learning and argued that joint learning and decoding may not necessarily give better results than individual approaches. \citet{NLP-scratch} also questioned the availability of fully labeled datasets as a bottleneck of such approaches. As discussed earlier, models represented by distributed state variables (see Fig. \ref{fig:multi-label-existing}b-c) are quite flexible and can be trained on sequences with missing labels by simply adding/dropping links whenever necessary, depending on the variable (label node) availability. It is hard to tackle such data for models represented by the joint state variable (see Fig. \ref{fig:multi-label-existing}a), as it requires both label categories $y_{1,t}$, $y_{2,t}$ to form the joint variable $y_t$ for training at all $t$. While most multi-task sequence learning models have been designed for problems in Natural Language Processing (NLP), \citet{multilabel-sequence-action} presented a joint model to improve the performance of action recognition and pose estimation from video data.

As elaborated in Section \ref{sec:intro}, we propose FLDCRF-m in order to model latent-dynamic interactions among different label categories. We describe how FLDCRF-m gives a generic expression for DCRFs with embedded hidden-state variables in Section \ref{sec:Models}. FLDCRF-m outperforms all state-of-the-art single (CRF, LDCRF, LSTM, LSTM-CRF) and multi-label (FCRF, CCRF, LSTM-m) models in the multi-label sequence tagging experiment.

Most studies on single and joint sequence labeling have been conducted in the context of NLP, viz., CoNLL 2000 chunking \citep{CoNLL-2000}, CoNLL 2003 named entity tagging \citep{CoNLL-2003} etc. However, recent reported (F1-measure) performance results on these tasks are largely driven by the quality of the input features, while the models do not show much variance in performance. By contrast, we wish to determine the effectiveness of the proposed models and therefore do not consider such datasets for evaluating our models. We plan to apply FLDCRF on NLP tasks in future.

We apply the proposed FLDCRF-s and FLDCRF-m models to two problems: 

\begin{enumerate}
\item Continuous gesture recognition on the UCI gesture phase segmentation dataset \citep{gesture-phase-data}. The task concerns online segmentation of gestures from rest positions.
\item Continuous multi-action recognition on the UCI opportunity dataset \citep{UCI-oppor}. This task aims at continuous joint recognition of two action types (locomotion and a high-level activity, see Section \ref{sec:Dataset}).
\end{enumerate}

We refer to Sections \ref{sec:Dataset} and \ref{sec:Experiments} for detailed description of the datasets and experiments respectively. All experiments in this paper consider continuous online recognition problems, i.e., the inference tasks are fomulated in terms of $P(y_t \mid x_{1:t})$. We do not utilize future input features ($x_{t+1}$, $x_{t+2}$ etc.) and future label/state variables during modeling/testing in any of the models.


\section{Factored Latent-Dynamic Conditional Random Fields (FLDCRF)}  \label{sec:Models}

We describe the proposed FLDCRF model here. As described in Section \ref{sec:intro}, a fully labeled FLDCRF has three variants:

\begin{enumerate}
\item FLDCRF single label sequence model (FLDCRF-s, see Fig. \ref{fig:LDCRF-FLDCRFs}b),
\item FLDCRF multi label sequence model (FLDCRF-m, see Fig. \ref{fig:FLDCRF-mi}a), and
\item FLDCRF multi agent interaction model (FLDCRF-i, see Fig. \ref{fig:FLDCRF-mi}b).
\end{enumerate}

All 3 three models can be expressed via a similar mathematical representation. Since the FLDCRF-s and FLDCRF-i are special cases of the FLDCRF-m model, we describe the FLDCRF-m model for better generalization. We describe how we construct the FLDCRF model, give its generic expression and then introduce the simplified versions we apply in this paper.

\subsection{Model} \label{subsec:FLDCRF-model}

Fig. \ref{fig:FLDCRF-mi}a shows the graph structure for FLDCRF in multi-label classification problems (FLDCRF-m). We depict the model for $L=2$ different label categories $\{y_{1,t}\}_{t=1:T}$ and $\{y_{2,t}\}_{t=1:T}$, interacting with each other through their respective hidden layers $\{h_{1,t}\}_{t=1:T}$ and $\{h_{2,t}\}_{t=1:T}$. In the case of a FLDCRF single-label model (FLDCRF-s, see Fig. \ref{fig:LDCRF-FLDCRFs}b), the label category $\{y_t\}_{t=1:T}$ is associated to all hidden layers. 

FLDCRF-m model (see Fig. \ref{fig:FLDCRF-mi}a) can be easily extended to accommodate multiple hidden layers for each label category $\{y_{i,t}\}_{t=1:T}$, $i = 1:L$. However, for simplicity of model description, we assume only one hidden layer per label category in FLDCRF-m. This leads to a total $L$ hidden layers in the model described below. We also assume first-order Markov connections among different hidden layers while applying FLDCRF in this paper. Higher order connections are possible but avoided for simplicity. These extensions can be utilized in order to improve model performance at the cost of model complexity and more training data, however we leave them to user discretion.

Let, \textbf{x} = \{\(x_1, x_2, ... , x_T\)\} denote the sequence of observations. \({\textbf{y}_i}\) = \{\(y_{i,1}, y_{i,2}, ... , y_{i,T}\)\} are the observed labels along layer $i$, $i = 1:L$. In the case of single-label prediction task by a FLDCRF-s (see Fig. \ref{fig:LDCRF-FLDCRFs}b), all layers take the same labels during model training, i.e., $y_{i,t}$ is same $\forall i$. In the case of multi-agent interaction model (FLDCRF-i, see Fig. \ref{fig:FLDCRF-mi}b), the features $x_t$ are distributed across $L$ layers, i.e., $x_t = \{x_{i,t}\}_{i = 1:L}$, assuming $L$ interacting agents in the environment. 

Let, \(\Upsilon_i \) be the alphabet for all possible label categories in layer $i$, $i = 1:L$, i.e., $y_{i,t}$ $\in$ $\Upsilon_i$. \({\textbf{h}_i}\) = \{\(h_{i,1}, h_{i.2}, ... , h_{i,T}\)\} constitutes the $i$-th hidden layer. Each possible class label \(\ell_i\) \(\in \) \(\Upsilon_i \) in layer $i$ is associated with a set of hidden states \(\textit{$\mathcal{H}$}_{i,\ell_i} \). \({\mathcal{H}_i}\) is the set of all possible hidden states for layer \textit{i} given by \({\mathcal{H}_i}\) = \(\bigcup_{\ell_i} {\mathcal{H}_{i,\ell_i}} \). 

The joint conditional model is defined as:

\begin{equation}
\begin{split}
\label{feqn1}
P\Big( \{{\textbf{y}_i}\}_{1:L} \mid \textbf{x}, \theta \Big) = \sum_{\{{\textbf{h}_i}\}_{1:L}} P\Big(\{{\textbf{y}_i}\}_{1:L} \mid \{{\textbf{h}_i}\}_{1:L}, \textbf{x}, \theta\Big) \cdot \; P\Big(\{{\textbf{h}_i}\}_{1:L} \mid \textbf{x}, \theta\Big).
\end{split}
\end{equation}

\begin{figure}[h]
\includegraphics[scale=0.6]{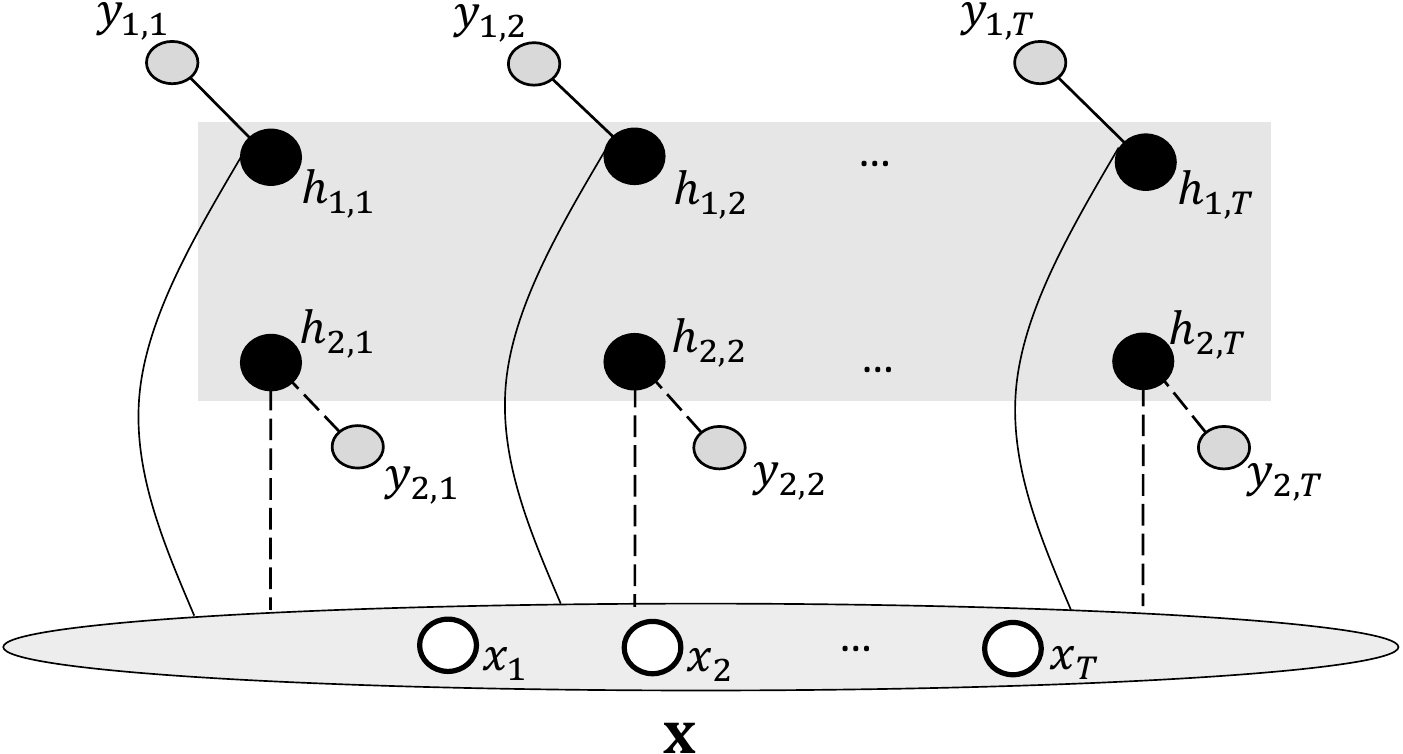}
\centering
\caption{The general structure of FLDCRF with the essential graphical constraints. Hidden variables $h_{i,t}$, $i = 1:L$, $t= 1:T$ can assume any dependency structure (first-order Markov, second-order Markov, cotemporal links between layers etc.) within themselves in this general FLDCRF structure. }
\label{fig:FLDCRF-gen}
\end{figure}

In order to keep model computations during training and testing tractable, we introduce the layers of hidden variables $\{{\textbf{h}_i}\}_{1:L}$ = $\{h_{i,t}\}_{i = 1:L, t=1:T}$ to the model with links (graphical constraints) as depicted in Fig. \ref{fig:FLDCRF-gen}. This allows us to factorize $P\Big(\{{\textbf{y}_i}\}_{1:L} \mid \{{\textbf{h}_i}\}_{1:L}, \textbf{x}, \theta\Big)$ according to equation \eqref{feqn1-explanation}.

\begin{equation}
\label{feqn1-explanation}
P\Big(\{{\textbf{y}_i}\}_{1:L} \mid \{{\textbf{h}_i}\}_{1:L}, \textbf{x}, \theta\Big) = \prod_{i=1}^{L}\prod_{t = 1}^{T} P({y_{i,t}} \mid {h_{i,t}}).
\end{equation}

Replacing $P\Big(\{{\textbf{y}_i}\}_{1:L} \mid \{{\textbf{h}_i}\}_{1:L}, \textbf{x}, \theta\Big)$ from equation \eqref{feqn1-explanation}, we can re-write equation \eqref{feqn1} as:

\begin{equation}
\begin{aligned}
\label{feqn1-new}
P\Big(\{{\textbf{y}_i}\}_{1:L} \mid \textbf{x}, \theta \Big) &= \sum_{\{{\textbf{h}_i}\}_{1:L} : \forall h_{i,t} \in \textit{$\mathcal{H}$}_{i,{y_{i,t}}}} \Bigg(\prod_{i=1}^{L}\prod_{t = 1}^{T} P({y_{i,t}} \mid {h_{i,t}})\Bigg) \cdot P\Big(\{{\textbf{h}_i}\}_{1:L} \mid \textbf{x}, \theta\Big) \\ 
&+ \sum_{\{{\textbf{h}_i}\}_{1:L} : \exists h_{i,t} \not\in \textit{$\mathcal{H}$}_{i,{y_{i,t}}}} \Bigg( \prod_{i=1}^{L}\prod_{t = 1}^{T} P({y_{i,t}} \mid {h_{i,t}})\Bigg) \cdot P\Big(\{{\textbf{h}_i}\}_{1:L} \mid \textbf{x}, \theta\Big).
\end{aligned}
\end{equation}

In order to further reduce model computations, we define the following model constraints, extending on \citet{LDCRF}:

\begin{enumerate}
\item{\(\textit{$\mathcal{H}$}_{i,\ell_i} \) are disjoint $\forall \ell_i \in \Upsilon_i $, $\forall i = 1:L$. }

\item{$h_{i,t}$ can only assume values from the set of hidden states assigned to the label $y_{i,t}$, i.e., $h_{i,t}$ $\in$ $\textit{$\mathcal{H}$}_{i,y_{i,t}}$, $\forall i = 1:L$ and $\forall t = 1:T$.}
\end{enumerate} 

These constraints let us write the following:

\begin{equation}
P({y_{i,t}}=\ell_i \mid {h_{i,t}}) = 
\begin{cases}
1, & h_{i,t} \in \mathcal{H}_{i,{y_{i,t}=\ell_i}} \\
0, & h_{i,t} \not\in \mathcal{H}_{i,{y_{i,t}=\ell_i}}.
\end{cases}
\label{feqn1-constraint}
\end{equation}

Thus, the FLDCRF model in equation \eqref{feqn1-new} can be reduced to:

\begin{equation}
\label{feqn3}
P\Big( \{{\textbf{y}_i}\}_{1:L} \mid \textbf{x}, \theta \Big) = \sum_{\{{\textbf{h}_i}\}_{1:L} : \forall h_{i,t} \in \textit{$\mathcal{H}$}_{i,{y_{i,t}}}} P\Big(\{{\textbf{h}_i}\}_{1:L} \mid \textbf{x}, \theta\Big).
\end{equation}

Equation \eqref{feqn3} gives the general expression of a FLDCRF. \\

\begin{figure}[h]
\includegraphics[scale=0.55]{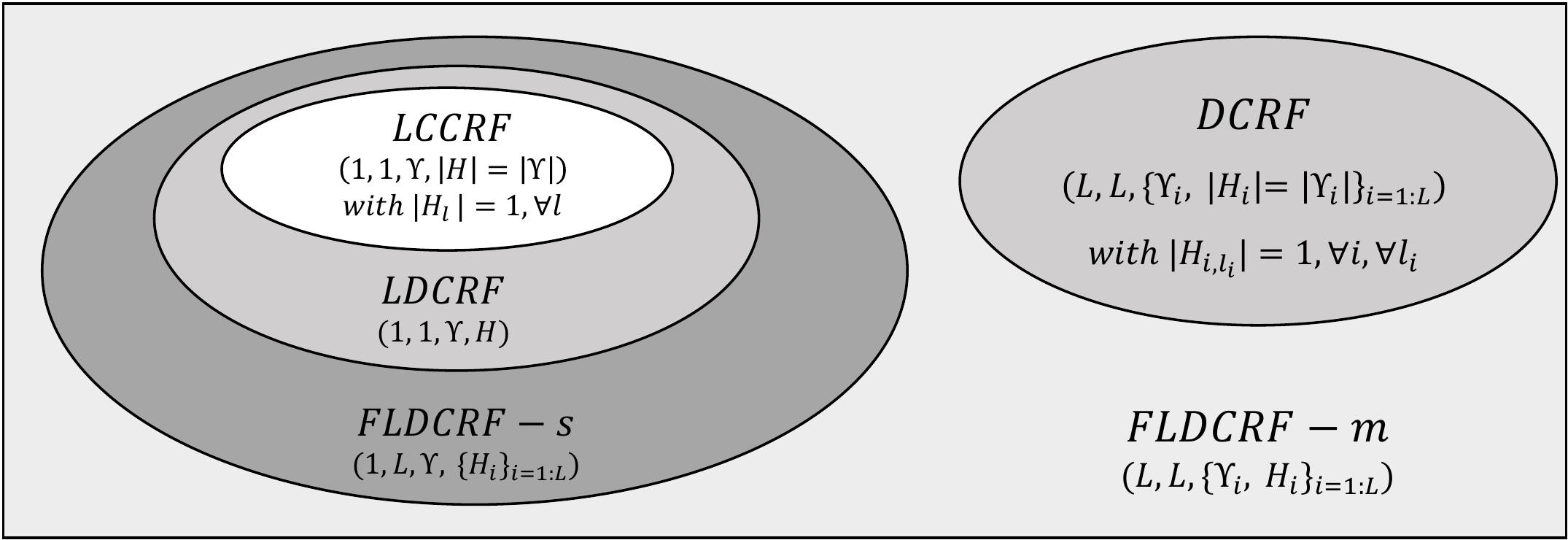}
\centering
\caption{Venn diagram illustrating relationship between FLDCRF and major sequential CRF variants. The configuration underneath the model names describe (number of label categories, number of hidden layers, class alphabet and the sets of hidden states) in a FLDCRF-m model in order to derive it. For simplicity of description, we assume 1 hidden layer for each label category in FLDCRF-m configuration, i.e., $L$ hidden layers for $L$ label categories.}
\label{fig:CRF-family}
\end{figure}

Equation \eqref{feqn3} simplifies to a LDCRF \citep{LDCRF} model for $L=1$. If we assume only one distinct hidden state per class label in each layer $i$, i.e., $\mid\mathcal{H}_{i,{l_i}} \mid \hspace{1mm}= \hspace{1mm}1$, $\forall i$, $\forall l_i$; and $\mid\mathcal{H}_{i} \mid \hspace{1mm}= \hspace{1mm} \mid \Upsilon_i \mid$, $\forall i$; equation \eqref{feqn3} simplifies to the joint conditional distribution $P\Big( \{{\textbf{y}_i}\}_{1:L} \mid \textbf{x}, \theta \Big)$ given by a DCRF \citep{DCRF} or a coupled CRF \citep{multilabel-sequence-3}. It is also straightforward to see that FLDCRF yields a LCCRF \citep{CRF} when $L=1$ and one distinct hidden state is associated to each class label. Thus, FLDCRF subsumes major sequential CRF variants viz., LDCRF, DCRF and LCCRF (illustrated in Fig. \ref{fig:CRF-family}).\\

We define $P\Big(\{{\textbf{h}_i}\}_{1:L} \mid \textbf{x}, \theta\Big)$ by the standard CRF formulation,

\begin{equation}
\label{feqn4}
P\Big(\{{\textbf{h}_i}\}_{1:L} \mid \textbf{x}, \theta\Big) = \frac{1}{\textbf{\textit{Z}}(\textbf{x},\theta)} \exp\left(\sum_k \theta_k . F_k\Big(\{{\textbf{h}_i}\}_{1:L},\textbf{x}\Big)\right),
\end{equation}

\noindent where index $k$ ranges over all parameters $\theta = \{\theta_k\}$ and $\textbf{\textit{Z}}(\textbf{x},\theta)$ is the partition function given by:

\begin{equation}
\label{feqn5}
\textbf{\textit{Z}}(\textbf{x},\theta) = \sum_{\{{\textbf{h}_i}\}_{1:L}} {\exp\left(\sum_k \theta_k . F_k\Big(\{{\textbf{h}_i}\}_{1:L},\textbf{x}\Big)\right)}.
\end{equation}

Now we introduce first-order Markov assumptions (as described earlier) among the hidden variables $\{\textbf{h}_{i}\}_{1:L}$. Therefore, the feature functions \(F_k \)'s are factorized (i.e., summed inside exponent) as: 

\begin{equation}
\label{feqn6}
F_k\Big(\{{\textbf{h}_i}\}_{1:L},\textbf{x}\Big) = \sum_{t=1}^T f_k\Big(\{h_{i,t-1}\}_{1:L}, \{h_{i,t}\}_{1:L}, \textbf{x}, \textit{t}\Big).
\end{equation}

We further factorize each $f_k\Big(\{h_{i,t-1}\}_{1:L}, \{h_{i,t}\}_{1:L}, \textbf{x}, \textit{t}\Big)$ at slice $t$, such that each hidden layer assumes its own (temporal) dynamics over the observations, while interacting with each other through cotemporal and first-order Markov (omitted in figures to avoid clutter) \textit{influence} links. Thus, each factorized component of $f_k\big(\{h_{i,t-1}\}_{1:L}, \{h_{i,t}\}_{1:L}, \textbf{x}, \textit{t}\big)$ at slice $t$ can be a \textit{state} function $s_k(h_{i,t}, \textbf{x}, t) $, a \textit{transition} function $t_k(h_{i,t-1},$ $h_{i,t}, \textbf{x}, t) $ or an \textit{influence} function $i_k\big(h_{i,t-l}, h_{j,t-m}, \textbf{x}, \textit{t}\big)$, with $i,j \in \{1:L\}$ and $(l,m) \in \{(0,0),(0,1),(1,0)\}$. We define \textit{state} and \textit{transition} functions by the following indicator functions:
\vspace{1mm}
\begin{equation}
\begin{split}
\label{eqn-state_tran}
s_k(h_{i, t}, \textbf{x}, \textit{t}) &= \mathbbm{1}_{\{(h_{i, t})= k\}}. x_t, \\
t_k(h_{i, t-1}, h_{i, t}, \textbf{x}, \textit{t}) &= \mathbbm{1}_{\{(h_{i, t},h_{i, t-1})= k\}},
\end{split}
\end{equation}
\vspace{1mm}
\noindent assuming one-hot representations for discrete $x_t$ components. We define \textit{influence} functions $i_k\big(h_{i,t-l}, h_{j,t-m}, \textbf{x}, \textit{t}\big)$, with $i,j \in \{1:L\}$ and $(l,m) \in \{(0,0),(0,1),(1,0)\}$ as:
\vspace{2mm}
\begin{equation}
\label{feqn-state_tran}
i_k\big(h_{i,t-l}, h_{j,t-m}, \textbf{x}, \textit{t}\big)= \mathbbm{1}_{\big\{\big(h_{i,t-l}, h_{j,t-m} \big)= k\big\}}.
\end{equation}

For the interaction model depicted in Fig. \ref{fig:FLDCRF-mi}b, the mathematical expressions are identical, only with a minor change in the \textit{state} function:

\begin{equation}
\label{eqn-state-interaction}
s_k(h_{i, t}, \textbf{x}_i, \textit{t}) = \mathbbm{1}_{\{(h_{i, t})= k\}}. x_{i,t} ,
\end{equation}

\noindent where $\textbf{x}_i = \{x_{i,t}\}_{t=1:T}$.\\

\subsection{Training}  \label{FLDCRF-training}

We estimate the model parameters by maximizing the conditional log-likelihood of the training data:
\vspace{2mm}
\begin{equation}
\label{feqn7}
\textit{\textbf{L}}(\theta) = \sum_{n=1}^N {\log P\left(\{{\textbf{y}_i}\}_{1:L}^{\left(n\right)} \mid \textbf{x}^{\left(n\right)}, \theta\right)} - \frac{{\parallel\theta\parallel}^2}{2\sigma^2},
\end{equation}

\noindent where \textit{N} is the total number of available labeled sequences. A L2 regularizer (second term in equation \eqref{feqn7}) was included in order to reduce overfitting during our experiments.

$P\left(\{{\textbf{y}_i}\}_{1:L}^{\left(n\right)} \mid \textbf{x}^{\left(n\right)}, \theta\right)$, $n = 1:N$ are obtained from equations \eqref{feqn3}-\eqref{feqn-state_tran}. The numerator $\textbf{\textit{N}}(\textbf{x}, \theta) = \sum_{\{{\textbf{h}_i}\}_{1:L} : \forall h_{i,t} \in \textit{$\mathcal{H}$}_{i,{y_{i,t}}}} \exp\left(\sum_k \theta_k . F_k\Big(\{{\textbf{h}_i}\}_{1:L},\textbf{x}\Big)\right)$ and the denominator $\textbf{\textit{Z}}(\textbf{x}, \theta)$ of equation \eqref{feqn3}, given by \eqref{feqn4} and \eqref{feqn5}, are in the classic sum-product form of dynamic programming and are efficiently computed by the forward algorithm \citep{HMM}. We apply the default BFGS optimizer in Stan modeling language \citep{Stan} to obtain the estimates $\hat{\theta} = \{\hat{\theta_k}\}$.

\subsection{Inference}  \label{FLDCRF-inference}

Multiple label sequences \({\textbf{y}_i}\), $i = 1:L$, can be inferred from the same graph structure by marginalizing over other labels:
\vspace{1mm}
\begin{equation}
\label{feqn8}
\hat{\textbf{y}}_i = \text{argmax}_{\textbf{y}_i} \quad \sum_{\big\{\{{\textbf{y}_i}\}_{1:L} - \; \textbf{y}_i\big\}} P\left(\{{\textbf{y}_i}\}_{1:L} \mid \textbf{x}, \hat{\theta}\right),
\end{equation}

\noindent where $\textbf{x} = \{x_t\}_{t = 1:T}$ is the observed input sequence of length $T$. $P\left(\{{\textbf{y}_i}\}_{1:L} \mid \textbf{x}, \hat{\theta}\right)$ can be obtained from \eqref{feqn3} and estimated parameters $\hat{\theta}$. At each instant $t$, the marginals $P(\{h_{i,t}\}_{1:L} \mid \textbf{x}, \hat{\theta})$ are computed and summed according to the disjoint sets of hidden states to obtain joint estimates of desired labels $\hat{y}_{i,t}$, $t = 1,2, ...$, $\forall i = 1:L $, as follows: 
\vspace{2mm}
\begin{equation}
\label{feqn9}
P(\{y_{i,t}\}_{1:L} \mid \textbf{x}) = \sum_{\{h_{i,t}\}_{1:L}:h_{i,t} \in \textit{$\mathcal{H}$}_{i,{y_{i,t}} }} P\left(\{h_{i,t}\}_{1:L} \mid \textbf{x}, \hat{\theta}\right).
\end{equation}

After marginalizing according to \eqref{feqn8}, the label $\hat{y}_{i,t}$ corresponding to the maximum probability is inferred. Since we consider online inference (i.e., input sequence observed upto current instant $t$) for our continuous gesture and action recognition problems, the inference problem in equation \eqref{feqn9} gets reduced to $P(\{y_{i,t}\}_{1:L} \mid {x_{1:t}})$. We compute the necessary probabilities $P\left(\{h_{i,t}\}_{1:L} \mid {x_{1:t}}, \hat{\theta}\right)$ by the forward algorithm \citep{HMM}. Forward-backward algorithm \citep{HMM} and Viterbi algorithm \citep{Viterbi} can also be applied for problems where online inference is not necessary.

\section{Datasets}  \label{sec:Dataset}

We apply our models on two datasets: a) UCI gesture phase segmentation dataset \citep{gesture-phase-data}, and b) UCI opportunity dataset \citep{UCI-oppor}. We describe the datasets below.

\subsection{UCI Gesture Phase Dataset}    \label{subsec:gesture-phase}

The UCI gesture phase data consists of features extracted from 7 videos with people gesticulating while telling stories. The data is captured by a Microsoft Xbox Kinect. Each time point within the 7 sequences is described by 18 positional features, 32 dynamic features (velocity, accleration etc.) and is labeled with one of the five gesture classes: rest, preparation, hold, stroke and retraction. The number of instances in each sequence is tabulated in Table \ref{tab:UCI-sequence-details}. `A', `B', and `C' denote different participants and `1', `2', `3' describe different stories. Although the data is multiclass, the most popular studies on the dataset \citep{gesture-phase-data, gesture-phase-appl1} have only considered binary classification, i.e., segmentation of gesture positions (i.e., positives - preparation, hold, stroke, retraction) from the rest (negative) position. We perform two binary time-series continuous gesture segmentation experiments on this dataset:

\begin{enumerate}
\item We follow the experiment 2 in \citep{gesture-phase-appl1}. The goal of the experiment is to train on the sequence A1 and test on sequence A2. We utilize the `Data vector 2' in the paper as our input features $x_t$, which consists of 3 dimensional positions of the left hand, right hand, left wrist and right wrist together, yielding a 12 dimensional input feature set. During experiment, we selected our models (hyperparameters, see Section \ref{sec:Experiments}) on a validation set comprising of the last 30\% frames of sequence A1, while training on the first 70\%. During testing, we re-trained the selected models on the entire A1 sequence.
\item A nested cross validation, with 7 outer loops (one for testing each sequence) and 6 inner loops (leave-one-out) per outer loop. We present results of both our experiments on the UCI gesture phase data in Section \ref{subsec:UCI-results}.
\end{enumerate}

\begin{table}[h]
\caption{Table showing number of instances in each sequence in UCI gesture phase dataset \citep{gesture-phase-data}.}
\label{tab:UCI-sequence-details}
\begin{center}
\begin{tabular}{| c| c| c| c| c| c| c| c| c|} 
\hline
Sequence & A1 & A2 & A3 & B1 & B3 & C1 & C3 & Total \\ 
\hline \hline
Num frames & 1747 & 1264 & 1834 & 1073 & 1423 & 1111 & 1448 & 9900 \\ 
\hline
\end{tabular}
\end{center}
\end{table}

\subsection{UCI Opportunity Dataset} \label{subsec:opportunity-data}

The UCI opportunity dataset \citep{UCI-oppor} is a public activity recognition dataset. The data contains 20 ADL (activity of daily living) sequences from 4 different participants (S1, S2, S3 and S4). Each participant has 5 ADL sequences. Each instance in the data has 3 activity label categories: 

\begin{itemize}
\item locomotion, which has four classes, viz., `lie', `sit', `stand' and `walk';
\item a high-level activity (HL), which has five classes, viz., `relaxing', `early morning', `coffee time', `sandwich time' and `clean up'; and
\item a mid-level activity (ML), which has several classes. 
\end{itemize}

\begin{figure}[h]
\includegraphics[scale=0.55]{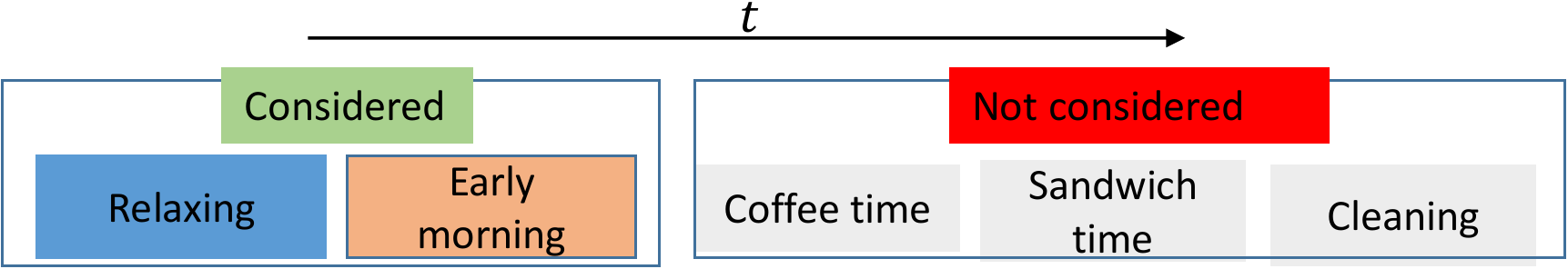}
\centering
\caption{Illustrating each ADL sequence in the UCI opportunity dataset \citep{UCI-oppor}.}
\label{uci-oppor-outline}
\end{figure}

We do not utilize the mid-level activity labels in this paper. The 5 high-level activities are performed in sequence within each complete ADL run (see Fig. \ref{uci-oppor-outline}). We consider the 5 sequences (ADL runs) from participant S2 in our experiment. From each ADL sequence, we utilize the data upto two high-level activity classes (i.e., `Relaxing' and `Early morning'), illustrated in Fig. \ref{uci-oppor-outline}. The `Relaxing' portions of the sequence primarily consists of the `lie' and `sit' locomotion classes, while the `early morning' activity class comes mostly with `stand' and `walk' locomotion classes, making the two label categories depeding on each other. Each time-point in the data also contains 242 sensor outputs (113 body-worn sensors, 32 shoe sensors and 97 other object and ambient sensors). We utilize the body-worn and shoe sensor outputs as our input features, yielding a 145 dimensional input feature set. For our convenience, we further divided the data into 28 sub-sequences, each fully labeled by the two label categories. Additionally, there are instances within the data with missing values for some sensor outputs. We replaced such values with the previous instant, if available, otherwise by `0's. Since the FLDCRF models encode the input features ($x_t$) by exponential functions over ($\theta_k \cdot x_t$), replacing values by `0's means a multiplication by 1, thus not affecting the model likelihood. We preprocessed each input dimension of $x_t$ between 0 to 1 for faster learning of our models.

We perform a nested cross-validation with 5 outer loops (with test sequences coming from each ADL sequence) and 4 inner loops (leave-one-ADL sequence out) per outer loop. Tabulated data (features and labels) for our multi-label sequence tagging experiment on UCI opportunity data is available here: https://github.com/satyajitneogiju/FLDCRF-for-sequence-labeling. Table \ref{tab:UCI-oppor-sequence-details} provides the details (length, ADL run etc.) of the 28 sub-sequences considered in this experiment.

Important dataset characteristics of the two datasets are listed in Table \ref{tab:dataset-compare}.

\begin{table}[h]
\caption{Number of instances in each sequence in the nested CV experiment on UCI opportunity dataset \citep{UCI-oppor}.}
\label{tab:UCI-oppor-sequence-details}
\begin{center}
\begin{tabular}{| c| c| c| c| c| c| c| c| c| c| c| c|} 
\hline
\backslashbox{\shortstack{Nested \\ outer set} }{Sequence} & 1 & 2 & 3 & 4 & 5 & 6 & 7 & 8 & 9 & Total\\ 
\hline \hline
1 & 1000 & 1000 & 800 & 943 & 724 & 847  & 1352 &  1276  & 298 & 8240 \\
\hline
2 & 721 &  863 &  800 &  1428 & 216 &  &  &  &  & 4028 \\
\hline
3 & 583 &  843 &  800 &  746  & 655 &  &  &  &  & 3627 \\
\hline
4 & 477 & 253 & 1008 &  954 & 583 &  &  &  &  & 3275 \\
\hline
5 & 711 & 1338  & 626  & 493  &  &  &  &  &  & 3168  \\
\hline
Total  &  &   &  &   &  &  &  &  &  & 22338 \\
\hline
\end{tabular}
\end{center}
\end{table}

\begin{table}[h]
\caption{Information on the UCI gesture phase \citep{gesture-phase-data} and the UCI opportunity dataset \citep{UCI-oppor}.}
\label{tab:dataset-compare}
\begin{center}
\begin{tabular}{| >{\centering\arraybackslash} m{3.5cm}| >{\centering\arraybackslash} m{6cm}| >{\centering\arraybackslash} m{6cm}|} 
\hline
\backslashbox{Attribute}{Dataset} & UCI gesture phase & UCI opportunity \\ [0.4ex] 
\hline \hline
Type & Single label & Multi label \\ [0.4ex] 
\hline
No. sequences & 7 & 5 \\ [0.4ex] 
\hline
No. sub-sequences & 7 & 28 \\ [0.4ex] 
\hline
Experiments & Experiment 2 \citep{gesture-phase-appl1}  Nested LOOCV & Nested LOOCV \\ [0.4ex] 
\hline
No. labels & 1 (Gesture) & 2 (Locomotion \& High-level activity) \\ [0.4ex] 
\hline
No. classes/label & 2 & 4 \& 2 \\ [0.4ex] 
\hline
Tested models & \shortstack{CRF, LDCRF, FLDCRF-s,\\ LSTM, LSTM-CRF}  & CRF, LDCRF, FLDCRF-s, LSTM, LSTM-CRF, FCRF, CCRF, FLDCRF-m, LSTM-m \\ [0.4ex] 
\hline
Input dimension ($|x_t|$) & 12 & 145 \\ [0.4ex] 
\hline
Metrics & F1-score & Micro F1-score \\ [0.4ex] 
\hline
\end{tabular}
\end{center}
\end{table}

\section{Experimental Setup}  \label{sec:Experiments}

\subsection{Metrics for evaluation}

We assess the different models by the F1 score. For the gesture phase segmentation problem on UCI gesture phase dataset, we consider rest position as `negative' and other gestures as `positive'. The F1 measure ($F$) is computed as:

\begin{equation}
F = \frac{2 \cdot P \cdot R}{P+R},
\label{f-score-gesture}
\end{equation}

\noindent where P and R represent precision and recall, defined as $P = \frac{TP}{TP+FP}$ and $R = \frac{TP}{TP+FN}$ respectively. $TP$, $FP$ and $FN$ are the predicted true positives, false positives and false negatives by the model.

For the multi-class action recognition problem on UCI opportunity data, we assess the models by the micro F1 score ($F_{micro}$), which also is defined by \eqref{f-score-gesture}, where precision and recall are given by:

\begin{equation}
P_{micro} = \frac{\sum_{c = 1}^{N_c}{TP_i}}{\sum_{c = 1}^{N_c}{TP_i}+\sum_{c = 1}^{N_c}{FP_i}}, 
\quad
R_{micro} = \frac{\sum_{c = 1}^{N_c}{TP_i}}{\sum_{c = 1}^{N_c}{TP_i}+\sum_{c = 1}^{N_c}{FN_i}}.
\label{prec-recall-multi}
\end{equation}

We apply the micro F1-score ($F_{micro}$) for evaluating the `locomotion' (4 classes) and overall (6 classes) performance. As both class labels (`relaxing' and `early morning') of the high-level activity (HL) are equally important, we slightly modify the precision ($P_{HL}$) and recall ($R_{HL}$) as follows:  
$P_{HL} = \frac{TP_{\text{early}} + TP_{\text{relax}}}{TP_{\text{early}} + TP_{\text{relax}}+FP_{\text{early}}}$ and $R_{HL} = \frac{TP_{\text{early}} + TP_{\text{relax}}}{TP_{\text{early}} + TP_{\text{relax}}+FN_{\text{early}}}$.

During validation, we select the multi-label sequential models (FLDCRF-m and LSTM-m) based on their combined performance (on all 6 classes) in the inner loops. The single-label models are separately selected on each label category.

\subsection{Benchmarking}  \label{subsec:Models}

For the single label sequence tagging tasks (`experiment 2' and nested CV) on the UCI gesture phase data, we compare CRF, LDCRF, FLDCRF-s, LSTM, and LSTM-CRF. For the multi-label sequence tagging task, we compare CRF, LDCRF, FLDCRF-s, LSTM, LSTM-CRF, FCRF, CCRF, FLDCRF-m, and LSTM-m. 

A FLDCRF-s model has two hyperparameters: number of hidden layers ($N_h$) and number of hidden states per label ($\{N_{si}\}$) along layers $i= 1:L$. For simplicity, we apply the same number of hidden states along all layers, i.e., $N_{si}= N_s$, $\forall i = 1:L$. We denote such model by FLDCRF-s($<$$N_h$$>$/$<$$N_s$$>$). For example, if we consider 2 hidden layers and 3 hidden states (per class label) across each layer, then the FLDCRF-s model will be denoted by FLDCRF-s($2/3$). In Table \ref{model-hyperpara}, we provide the list of tested hyperparameter settings of FLDCRF-s across the 3 different experiments.

\begin{figure}[h]%
\centering
\subfloat[\scriptsize{FLDCRF-m1.}]{{\includegraphics[scale=0.5]{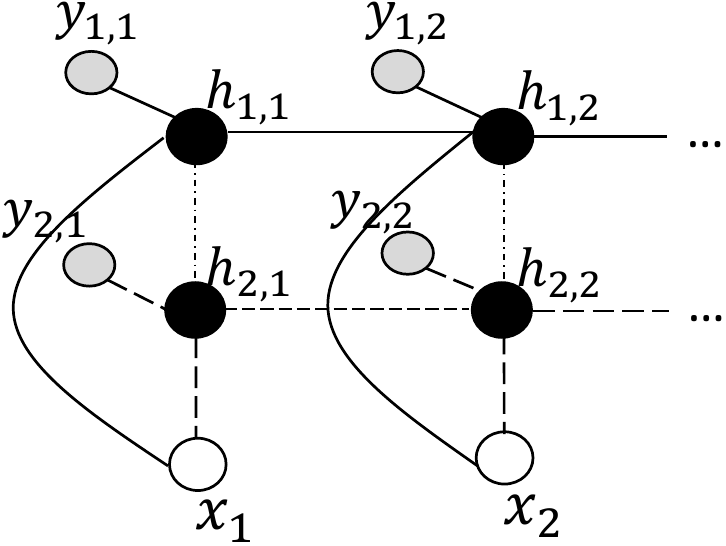} }}%
\hspace{1.5em}
\subfloat[\scriptsize{Cliques at $t$ for FLDCRF-m1.}]{{\includegraphics[scale=0.5]{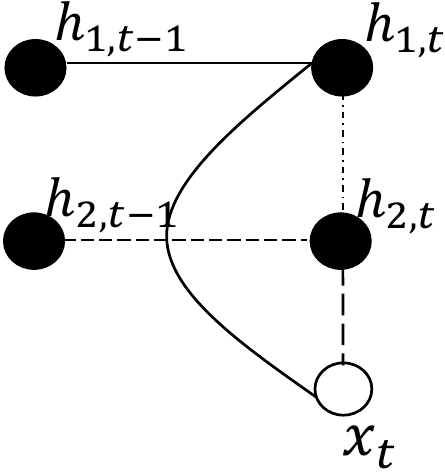} }}%
\hspace{3em}
\subfloat[\scriptsize{FLDCRF-m2.}]{{\includegraphics[scale=0.5]{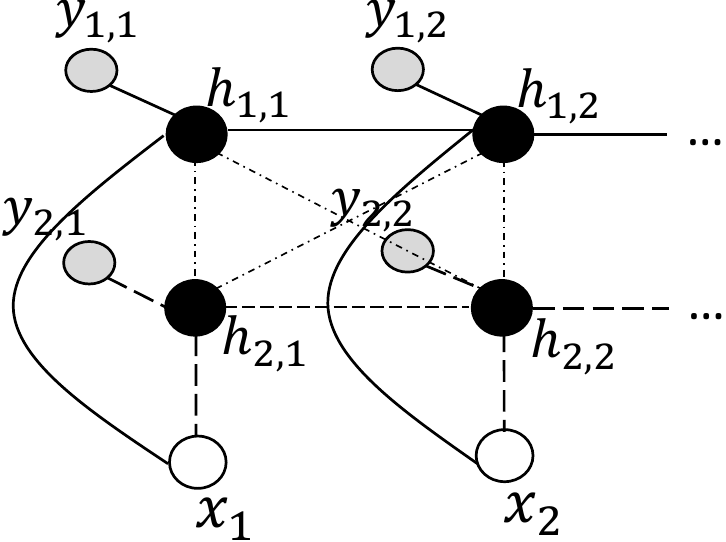} }}%
\hspace{1.5em}
\subfloat[\scriptsize{Cliques at $t$ for FLDCRF-m2.}]{{\includegraphics[scale=0.5]{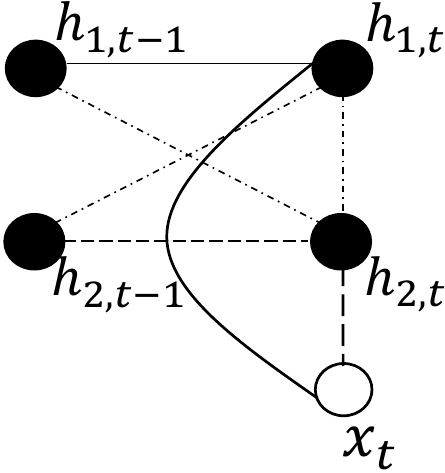} }}%
\caption{Two different FLDCRF-m models considered in the experiment on the UCI opportunity data.}%
\label{fig:FLDCRF-m1m2}%
\end{figure}

A FLDCRF-m (multi-label) model has three hyper-parameters: number of different label categories $N_l$, number of hidden layers per label category $\{N_{hl}\}_{l=1:N_l}$, and number of hidden states per label $\{N_{si}\}$ along layers $i= 1:L$, where $L = \sum_{l=1}^{N_l}{\{N_{hl}\}}$. For simplicity, we only keep one hidden layer ($N_{hl} = 1, \forall l=1:N_l$) for each label category, thus making total number of hidden layers equal to the number of different label categories, i.e., $L$ = $N_l$. We denote such a model by FLDCRF-m$\{N_{si}\}_{i= 1:L}$. For example, if there are 3 label categories and we associate 1, 2 and 4 hidden states (per class label) respectively to each label category, then the 3-layered FLDCRF-m model will be represented as FLDCRF-m$\{1, 2, 4\}$. In our multi-label experiment on the opportunity data, there are two different label categories (locomotion and high-level), and the FLDCRF-m models are denoted as FLDCRF-m$\{N_{s1}, n_{s2}\}$. In Table \ref{model-hyperpara}, we list the considered hyperparameter settings of FLDCRF-m. We considered two different FLDCRF-m models (with and without the first-order Markov influence among hidden layers, see Fig. \ref{fig:FLDCRF-m1m2}) during our multi-label experiment on the UCI opportunity data.

An LSTM has three hyperparameters: number of hidden units ($N_{hls}$), number of epochs to train ($N_{els}$), and training minibatch size ($N_{mls}$). To the best of our knowledge, there is no generally accepted rule to select the optimal $N_{hls}$, therefore, they need to be tuned by trial and error. As suggested in \citep{HU-LSTM-blog, thomas-ans-lstm, HU-size-lstm}, we vary the $N_{hls}$ according to,

\begin{equation}
N_{hls} = \frac{N_{sa}}{\alpha \cdot (N_i + N_o)},
\label{eq:select-lstm-model}
\end{equation}

\noindent where $N_{sa}$ is the number of samples in the training dataset (number of instances; see Tables \ref{tab:UCI-sequence-details} and \ref{tab:UCI-oppor-sequence-details}), $N_i$ is the number of input neurons ($|x_t|$ in our case; see Table \ref{tab:dataset-compare}) and $N_o$ is the number of output neurons from the LSTM layer. We vary the parameter $\alpha$ between 2 and 10 to select the $N_{hls}$'s. A few other popular suggestions include:

\begin{itemize}
\item Keep $N_{hls}$ between $N_i$ and $N_o$,
\item Set $N_{hls}$ as the arithmetic mean of $N_i$ and $N_o$,
\item Set $N_{hls}$ as the geometric mean of $N_i$ and $N_o$ etc.
\end{itemize}

We try to follow these recommendations as closely as possible while selecting $N_{hls}$ values to be tested. In Table \ref{model-hyperpara}, we list the considered LSTM $N_{hls}$ values in each of our experiments. During validation, we run each model setting ($N_{hls}$) for 500 epochs, saving model performances on validation sets at every 100 epochs. We feed all training sequences to the LSTM models at each epoch at the rate of 1 sequence per minibatch.

\begin{table}[h]
\caption{Hyperparameter settings for FLDCRF-s, FLDCRF-m, LSTM and LSTM-m.}
\label{model-hyperpara}
\begin{center}
\begin{tabular}{| >{\centering\arraybackslash} m{3.4cm}| >{\centering\arraybackslash} m{2.9cm}| >{\centering\arraybackslash} m{3.1cm}| >{\centering\arraybackslash} m{2cm}| >{\centering\arraybackslash} m{2cm}|} 
\hline 
\multirow{2}{*}{Experiment} & \multicolumn{4}{c|}{Models} \\
\cline{2-5}
\multicolumn{1}{|c|}{}  & FLDCRF-s ($N_h/N_s$) & FLDCRF-m ($\{N_{s1}, n_{s2}\}$) & LSTM ($N_{hls}$) & LSTM-m ($N_{hls}$) \\ [0.4ex] 
\hline \hline
\shortstack{Experiment 2 \\ \citep{gesture-phase-appl1}} & 1/1, 1/2, 1/3, 1/4, 1/5, 1/6, 1/7, 2/1, 2/2, 2/3, 2/4, 3/1, 3/2, 3/3 & NA & 2, 5, 7, 10, 14, 20, 50, 100, 200 & NA \\ [0.4ex] 
\hline
Nested CV on Gesture phase & 1/1, 1/2, 1/3, 1/4, 1/5, 1/6, 2/1, 2/2, 2/3, 2/4, 2/5, 3/1, 3/2, 3/3 & NA & 2, 5, 7, 10, 14, 20, 50, 100, 200 & NA\\ [0.4ex] 
\hline
Nested CV on Opportunity (Locomotion) & 1/1, 1/2, 1/3, 1/4, 2/1, 2/2, 2/3 & \multirow{2}{*}{\shortstack{\{1,1\}, \{1,2\}, \{1,3\},\\ \{1,4\}, \{1,5\}, \{1,6\},\\ \{2,1\}, \{2,2\}, \{2,3\},\\ \{2,4\}, \{2,5\}, \{2,6\},\\\{3,1\}, \{3,2\}, \{3,3\}}} & 5, 10, 25, 50, 75, 150, 300, 500 & \multirow{2}{*}{\shortstack{5, 10, 25, \\50, 75, 150,\\ 300, 500}}\\ [0.4ex] 
\cline{1-2}\cline{4-4}
Nested CV on Opportunity (HL) & 1/1, 1/2, 1/3, 1/4, 1/5, 1/6, 2/1, 2/2, 2/3, 2/4 & \multicolumn{1}{c|}{} & 2, 5, 10, 25, 50, 75, 150 & \multicolumn{1}{c|}{}\\ [0.4ex] 
\hline
\end{tabular}
\end{center}
\end{table}

\subsection{Model Platforms and System Specifications}

We trained and tested FLDCRF type models (FLDCRF, LDCRF, CRF, CCRF etc.) by the PyStan interface of Stan modeling language \citep{Stan}. Stan's in-built BFGS and L-BFGS optimizers allow us to conveniently train models by defining the model likelihoods. We apply BFGS to train all FLDCRF models and keep all default settings for the optimizer.

We implemented the LSTM models defined in Keras, a deep learning library for Python running Tensorflow in the backend. LSTM parameters were trained by an Adam optimizer with a default learning rate of 0.001 and other default parameters. To reduce overfitting, we added a dropout layer (with regularization 0.2) between the LSTM and softmax (or CRF for LSTM-CRF model) layers for all models with $\ge$5 hidden units. We also added a dropout (with regularization 0.2) to the LSTM layer for all models with $\ge$5 hidden units. Our Pystan FLDCRF and Keras LSTM implementation codes are available from - https://github.com/satyajitneogiju/FLDCRF-for-sequence-labeling.

We perform the metric computations and plot figures in MATLAB 2015b \citep{MATLAB}. The LSTM models are trained (and tested) on an Nvidia Tesla K80 GPU, while the FLDCRF models are trained (and tested) on an Intel(R) Xeon(R) CPU E5-2630 v3 @2.40GHz CPU.

\section{Results} \label{sec:Results}

A good model should not only perform well on the validation data, it must be consistent across validation and test data. At the same time, the model selection process should be lucid and easy. It is also desirable to have fast model training and inference mechanisms, preferably without additional resources, e.g., GPU. Therefore, we not only compare different models (FLDCRF, LSTM, LSTM-CRF etc.) on the test data, but also examine several other modeling aspects, viz., ease of model selection, consistency across validation and test performance, and computation times. We present the following 4 types of model performance measures for each experiment:

\begin{itemize}
\item \textit{Test performance:} Under this measure, different models are compared on the test data.
\item \textit{Ease of model selection:} Under this performance attribute, we discuss the model hyperparameter selection process of FLDCRF and LSTM families on the validation data. In order for this model selection process to be fast, easy and effective, a model must have the following characteristics: 

	\begin{enumerate}
	\item \underline{Fewer hyperparameters}: There should be very few types of hyperparameters which strongly influence the model performance. FLDCRF models only depend on the $N_h$/$N_s$ settings, and do not need to be tuned for number of training epochs. LSTM models, on the other hand, need to be optimized for number of training epochs alongside the number of hidden units $N_{hls}$ and training minibatch size $N_{mls}$. 
	\item \underline{Rule to choose hyperparameters:} It is best to have definite rules to choose the model hyperparameters, rather having to select them via trial and error. 
	\item \underline{Observable pattern in validation performance}:  In order to easily select the model hyperparameters, the model must show discernible patterns (increasing/decreasing) in validation performance against the choices of the hyperparameters. Model performance across different hyperparameter settings should not vary at random.
	\item \underline{Lesser variance:} In order to avoid tedious hyperparameter selection and continuous monitoring of model performance on validation data, model performance across adjacent hyperparameter settings should not vary widely. 

Wide and random variation across hyparameters (on validation data) without any pattern also require widespread selection of hyperparameters.
	\item \underline{Fair worst case performance:} The model must gurantee good performance with minimal effort to optimize hyperparameters. To accomplish this, worst case performances reported by the model on validation data should not differ largely from the best/average performances, as well as percentage of such poor outcomes should be very low (if any).
	\end{enumerate}
\item \textit{Consistency:} We examine consistency of the selected models on test data under this measure. Consistency in model performance across validation and test data reflects stability of a model for practical deployment.
\item \textit{Computation times:} In this result category, we present computation times required by the models for training and inference, for different choices of the hyperparameters.
\end{itemize}

\subsection{UCI Gesture Phase Dataset} \label{subsec:UCI-results}

As mentioned earlier, we perform two experiments (see Section \ref{subsec:gesture-phase}) on the UCI Gesture Phase Dataset for segmenting gestures from rest positions. In the first experiment, we follow the experiment 2 by \citet{gesture-phase-appl1}. In the second experiment, we perform a 7 (outer) fold nested cross-validation.

\subsubsection{Experiment 1: \citep{gesture-phase-appl1}}

\begin{itemize}[leftmargin=*]

\item \textit{Test performance:}

Table \ref{tab:experiment2-test} compares different models on the test sequence A2 of the UCI gesture phase dataset. FLDCRF-s considerably outperforms LSTM and LSTM-CRF models on the test set. FLDCRF-s also outshines a multi-layered perceptron (MLP) model \citep{gesture-phase-appl1} reported on the test data. However, FLDCRF-s does not improve LDCRF performance in this experiment. 

Considering variable performance on re-training the LSTM models, we also report performance of the optimized LSTM-CRF model that is not re-trained on the combined training and validation data. FLDCRF-s outperforms this LSTM-CRF model (F1 score 85\%) as well on the test data.

\begin{table}[h!]
\caption{F1 scores of the models for the experiment proposed by \citet{gesture-phase-appl1} on UCI gesture phase data. The best LSTM-CRF model on the validation set achieved a F1 score 84.88 without retraining on the training+validation set.}
\label{tab:experiment2-test}
\begin{center}
\begin{tabular}{| >{\centering\arraybackslash} m{3.5cm}| >{\centering\arraybackslash} m{3.5cm}|} 
\hline
Model & F1-score (\%) \\ [0.4ex] 
\hline
\hline
CRF & 81.35 \\ [0.4ex] 
\hline
LDCRF & \textbf{88.60} \\ [0.4ex] 
\hline
FLDCRF-s & \textbf{88.60} \\ [0.4ex] 
\hline
LSTM & 83.42 \\ [0.4ex] 
\hline
LSTM-CRF &  84.28 (84.88)\\ [0.4ex] 
\hline
Multi-layered Perceptron \citep{gesture-phase-appl1} & 82.34 \\ [0.4ex] 
\hline
\end{tabular}
\end{center}
\end{table}

\item \textit{Model selection:}

FLDCRF-s models only need to be optimized for number of hidden layers $N_h$ and number of hidden states $N_s$. Each LSTM model (LSTM and LSTM-CRF, with a given number of hidden units $N_{hls}$), on the other hand, need to be tuned for the optimum number of training epochs; thus requiring careful monitoring on validation performance in order to avoid overfitting. We analyze the selection process of the best performing models in each family, viz., FLDCRF-s and LSTM-CRF below.

Table \ref{tab:FLDCRF-validation-exp2} presents FLDCRF-s performance on validation data. The models continue to perform better on increasing $N_s$ upto 6 with $N_h = 1$, but decline on increasing $N_h$, giving clear indications to stop testing more hyperparameters. On the other hand, LSTM-CRF model performances vary rapidly across the training epochs (see Table \ref{tab:LSTM-validation-exp2}) for most of the $N_{hls}$ settings. While in some cases (e.g., $N_{hls}$ = 10, 14 etc.) model performance goes down rapidly after attaining the maximum, there are cases (e.g., $N_{hls}$ = 100, 200 etc.) where the validation F1-score varies randomly across training epochs, without any discernible pattern against the hyperparameter. Such random variation demands very careful monitoring over the validation outcomes. Even when optimized for the training epochs for each $N_{hls}$ setting (see column 7 of Table \ref{tab:LSTM-validation-exp2}), F1 scores do not reveal any pattern over the $N_{hls}$'s. Such a behaviour additionally brings in the problem of considering widespread values of $N_{hls}$.

\begin{table}[h!]
\caption{FLDCRF-s validation performance on experiment 2 \citep{gesture-phase-appl1} on UCI gesture phase data.}
\label{tab:FLDCRF-validation-exp2}
\begin{center}
\begin{tabular}{| c| >{\centering\arraybackslash} m{0.85cm}| >{\centering\arraybackslash} m{0.85cm}| >{\centering\arraybackslash} m{0.85cm}| >{\centering\arraybackslash} m{0.85cm}| >{\centering\arraybackslash} m{0.85cm}| >{\centering\arraybackslash} m{0.85cm}| >{\centering\arraybackslash} m{0.85cm}| >{\centering\arraybackslash} m{0.8cm}| >{\centering\arraybackslash} m{0.85cm}| >{\centering\arraybackslash} m{0.85cm}|} 
\hline
FLDCRF-s & 1/1 & 1/2 & 1/3 & 1/4 & 1/5 & 1/6 & 1/7 & Best & Worst & Std\\ 
\hline
F1 & 80.07 & 80.79 & 80.87 & 81.31 & 81.6 & \textbf{82.96} & 81.92  & \multirow{3}{*}{82.96} & \multirow{3}{*}{\textbf{76.47}} & \multirow{3}{*}{\textbf{2.11}}\\
\cline{1-8} \cline{1-8}
FLDCRF-s & 2/1 & 2/2 & 2/3 & 2/4 & 3/1 & 3/2 & 3/3  & \multicolumn{1}{c|}{} & \multicolumn{1}{c|}{} & \multicolumn{1}{c|}{}\\ 
\cline{1-8}
F1 & 77.57 & 77.84 & 78.97 & 78.03 & 76.84 & 76.60 & \textcolor{gray}{\textbf{76.47}}  & \multicolumn{1}{c|}{} & \multicolumn{1}{c|}{} & \multicolumn{1}{c|}{}\\
\hline
\end{tabular}
\end{center}
\end{table}

\begin{table}[h!]
\caption{LSTM-CRF validation performance on experiment 2 \citep{gesture-phase-appl1} on UCI gesture phase data. $N_{els}$ denotes number of training epochs and $N_{hls}$ represents the number of hidden units in the model.}
\label{tab:LSTM-validation-exp2}
\begin{center}
\begin{tabular}{| >{\centering\arraybackslash} m{1.3cm}| >{\centering\arraybackslash} m{0.62cm}| >{\centering\arraybackslash} m{0.62cm}| >{\centering\arraybackslash} m{0.62cm}| >{\centering\arraybackslash} m{0.62cm}| >{\centering\arraybackslash} m{0.62cm}| >{\centering\arraybackslash} m{0.62cm}| >{\centering\arraybackslash} m{0.62cm}| >{\centering\arraybackslash} m{0.62cm}| >{\centering\arraybackslash} m{0.62cm}| >{\centering\arraybackslash} m{0.85cm}| >{\centering\arraybackslash} m{1.5cm}| >{\centering\arraybackslash} m{0.65cm}|} 
\hline
\backslashbox{$N_{els}$\kern-0.8em}{\kern-0.8em$N_{hls}$} & 2 & 5 & 7 & 10 & 14 & 20 & 50 & 100 & 200 & Best & Worst & Std\\ 
\hline \hline
100 & 68.9 & 78.7 & 78.8 & \textbf{81.3} & 80 & 83.9 & 73.4 & 78.8 & 72.1 & \multirow{5}{*}{\textbf{87.17}} & \multirow{5}{*}{\shortstack{62.45 \\(18\%$<$75)}} & \multirow{5}{*}{5.59}\\
\cline{1-10}
200 & 75 & 81.1 & 82.5 & 71.9 & 75.6 & 76.5 & 81.4 & 68.2 & 86.4 & \multicolumn{1}{c|}{} & \multicolumn{1}{c|}{} & \multicolumn{1}{c|}{}\\ 
\cline{1-10}
300 & 79.9 & 80.1 & 83.6 & \textcolor{gray}{\textbf{67.4}} & \textbf{87.2} & 80.3 & 82.8 & 75.9 & 77.7 & \multicolumn{1}{c|}{} & \multicolumn{1}{c|}{} & \multicolumn{1}{c|}{}\\
\cline{1-10}
400 & 81.8 & 78.6 & 83.4 & \textcolor{gray}{\textbf{63.3}} & 80.2 & 79.1 & 77.3 & 75.3 & 84.7 & \multicolumn{1}{c|}{} & \multicolumn{1}{c|}{} & \multicolumn{1}{c|}{}\\
\cline{1-10}
500 & 83.4 & 79.4 & 84.2 & \textcolor{gray}{\textbf{62.5}} &  \textcolor{gray}{\textbf{76.5}} & 81.2 & 77.8 & 75.1 & 76.9 & \multicolumn{1}{c|}{} & \multicolumn{1}{c|}{} & \multicolumn{1}{c|}{}\\
\hline \hline
Best & 83.4    &     81.1    &     84.2     &    81.3     &    87.2     &    83.9     &    82.8     &    78.8    &     86.4  &  &  & \multicolumn{1}{c|}{}\\
\hline
Std & 5.9 & 1.01 & 2.17 & 7.69 & 4.54 & 2.71 & 3.7 & 3.9 & 5.88  &  &  & \multicolumn{1}{c|}{}\\
\hline
\end{tabular}
\end{center}
\end{table}

FLDCRF-s models do not vary rapidly (standard deviation 2.11) across the choices of the hyperparameters, and rather indicate a worst case F1-score of 76.47\%. On the other hand, a high percentage (18\%) of the LSTM-CRF models generate poor F1 performance ($<$75\%) with a notably low worst case performance (62\%). In addition, LSTM-CRF models display a considerably higher standard deviation (5.11) among validated models.  

Since there is no rule to select the best performing hyperparameter(s) for a LSTM-CRF, large and rapid variation in model performance across hyperparameter settings, high-percentage of poor performances with notably low worst cases, and longer training times (see below) make it very difficult for a fast and reliable LSTM model selection. We will see similar LSTM behaviour in all our experiments. \\

\item \textit{Consistency:} 

Although FLDCRF-s offers easier model selection than LSTM-CRF, LSTM-CRF outperforms FLDCRF-s on the best validation performance, yielding a F1-score 87\% compared to 83\% by FLDCRF-s. However, this optimized LSTM-CRF model, when re-trained with same hyperparameters on training+validation data, fails to outperform FLDCRF-s on the test data (see Table \ref{tab:experiment2-test}), achieving 84.3\% compared to 88.6\% by FLDCRF-s. Furthermore, considering variable performance for LSTM-CRF on re-training the model, we apply the best (pre-trained) LSTM-CRF model (with 87\% on validation) on the test data. However, this model (F1-score 84.9\%) also fails to outperform FLDCRF-s on the test data (see Table \ref{tab:experiment2-test}).

In addition to a tedious model selection process, LSTM-CRF displays inconsistent performance across validation and test data. This inconsistency is reflected throughout the experiments in the paper.

\item \textit{Computation times:}

We compare training and inference times required by the FLDCRF and LSTM models in Fig. \ref{fig:FLDCRF-time}. While few FLDCRF-s models and LSTM required similar computation times for inference, computation time required to train the FLDCRF-s models is notably lower than the LSTM models. Furthermore, FLDCRF computation times can be significantly reduced by GPU implementation and careful optimization, allowing more hidden layers ($N_h$) and states ($N_s$) for large datasets, if necessary. We consider the GPU implementation and application of FLDCRF on big data as our future work.

\end{itemize}

\begin{figure}[h!]%
\centering
\subfloat[\small{Training time.}]{{\includegraphics[scale=0.55]{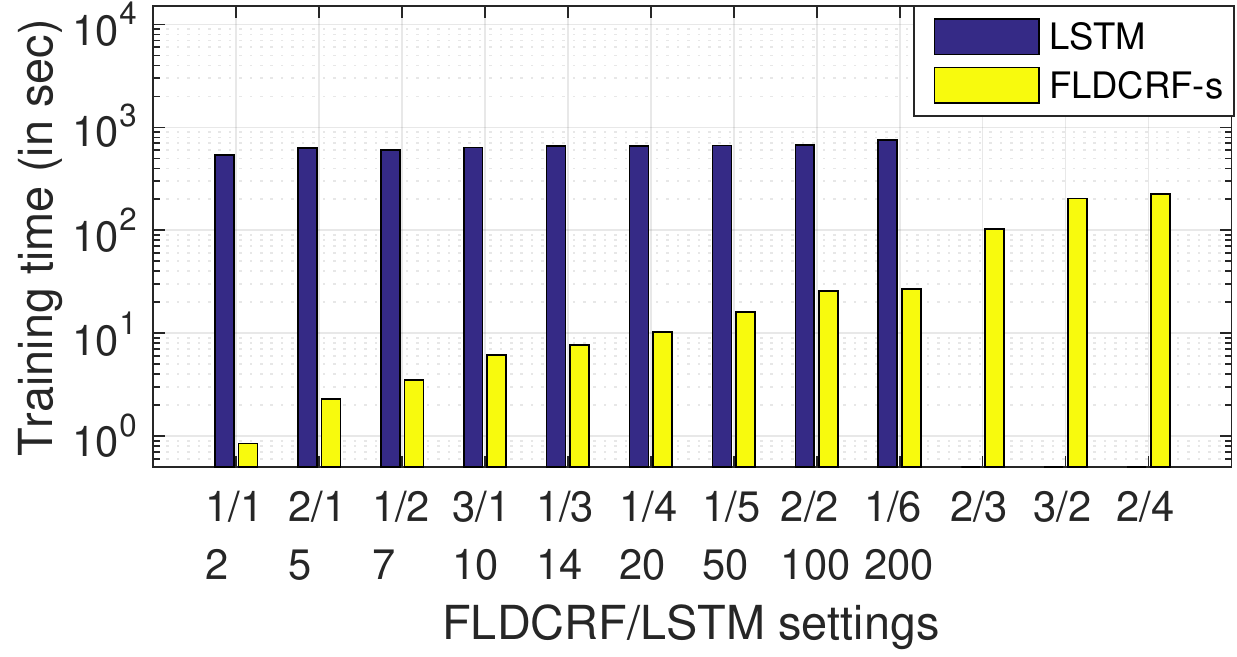} }}%
\hspace{2em}
\subfloat[\small{Average inference time per frame.}]{{\includegraphics[scale=0.55]{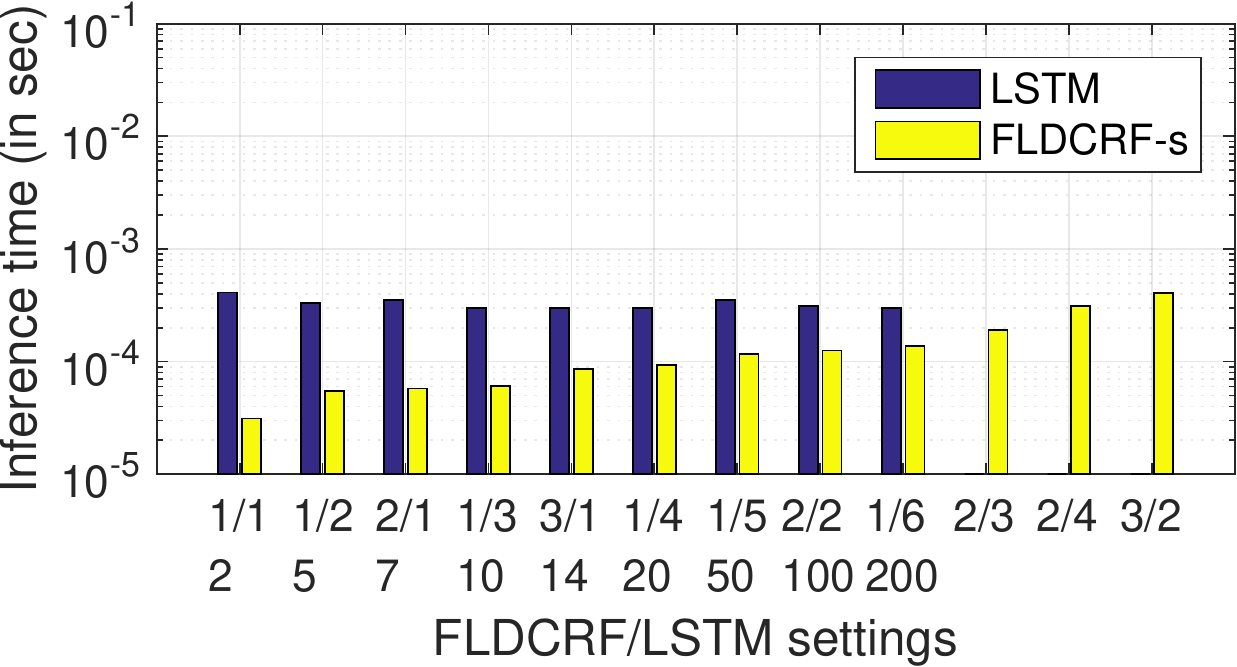} }}%
\caption{Training and inference times of different FLDCRF and LSTM models on UCI gesture phase data. Training time shown for each FLDCRF-s is until model convergence on training data, with the default criteria in Stan. Training time shown for LSTM is for 500 training epochs.}%
\label{fig:FLDCRF-time}%
\end{figure}

\subsubsection{Experiment 2: nested CV}

Since the training and test datasets in Experiment 1 were small, we consider the entire UCI gesture phase dataset in this experiment. We perform a nested cross-validation. Since there are 7 large sequences in the dataset (see Section \ref{subsec:gesture-phase}), we divide the data in 7 outer folds, one for testing each sequence. In each case, we select the models by a 6-fold (leave-one-out) cross-validation on the remaining sequences. \\

\begin{itemize}[leftmargin=*]
\item \textit{Test performance:} 

Table \ref{tab:nested-cv-test-gesture} compares different models on the test sets. FLDCRF-s boosts LDCRF performance on 4 sets (sets 1, 2, 3, 7), with notable improvements ($\sim$3\% and $\sim$2\%) on sets 1 and 3. As a result, FLDCRF-s improves the overall LDCRF performance to outshine the overall LSTM (and LSTM-CRF) performance on the test sets. FLDCRF-s individually outperforms LSTM and LSTM-CRF models on 4 (out of 7) sets each.

\begin{table}[h]
\caption{Different model performances on the nested CV experiment on UCI gesture phase data.}
\label{tab:nested-cv-test-gesture}
\begin{center}
\begin{tabular}{| >{\centering\arraybackslash} m{2.5cm}| >{\centering\arraybackslash} m{1cm}| >{\centering\arraybackslash} m{1cm} | >{\centering\arraybackslash} m{1cm}| >{\centering\arraybackslash} m{1cm}| >{\centering\arraybackslash} m{1cm}| >{\centering\arraybackslash} m{1cm}| >{\centering\arraybackslash} m{1cm}| >{\centering\arraybackslash} m{1.5cm}|} 
\hline
\multirow{2}{*}{Model} & \multicolumn{8}{c|}{F1 score} \\ [0.4ex] 
\cline{2-9}
\multicolumn{1}{|c|}{} & 1 & 2 & 3 & 4 & 5 & 6 & 7 & Average \\
\hline
LDCRF & 85.6 & 76.92 & 87.81 & 92.6 & \textbf{92.76} & \textbf{91.83} & 76.1 & 86.23 \\
\hline
FLDCRF-s & 88.4 & 76.95 & \textbf{89.69} & 92.6 & \textbf{92.76} & 91.38 & \textbf{76.17} & \textbf{86.85} \\
\hline
LSTM & 90.95 & \textbf{81.61} & 82.59 & \textbf{94.41} & 92.51 & 89.12 & 73.5 & 86.39 \\
\hline
LSTM-CRF & \textbf{91.99} & 81.51 & 81.60 & 93.40 & \textbf{92.76} & 86.87  & 74.06 & 86.03  \\
\hline
\end{tabular}
\end{center}
\end{table}

\item \textit{Model selection:} 

Tables \ref{tab:FLDCRF-validation-nested-gesture} and \ref{tab:LSTM-validation-nested-gesture} present FLDCRF-s and LSTM cross-validation F1 outcomes on sets 4 and 7 respectively. Similar to experiment 1, LSTM models display rapid variation in performance across training epochs, demanding careful tracking of validation performance. Few such instances are highlighted in Table \ref{tab:LSTM-validation-nested-gesture}. FLDCRF-s does not require to tune the training epochs, and variation among adjacent hyperparameters (see Table \ref{tab:FLDCRF-validation-nested-gesture}) is neither rapid nor random, with observable decay in performance due to overfitting beyond certain settings of the hyperparameters. 

\begin{table}[h]
\caption{FLDCRF-s validation performance on nested validation sets 4 and 7 of the nested CV experiment on UCI gesture phase data.}
\label{tab:FLDCRF-validation-nested-gesture}
\begin{center}
\begin{tabular}{| c| c| c| c| c| c| c| c| c| c|} 
\hline
\multicolumn{10}{|c|}{Set 4} \\
\hline \hline
FLDCRF & 1/1 & 1/2 & 1/3 & 1/4 & 1/5 & 1/6 & Best & Worst  & Std\\ 
\hline
F1 & 86.38 & 85.58 & \textbf{86.73} & 86.5 & 86.33 & 85.96 & \multirow{3}{*}{86.73} & \multirow{3}{*}{\textbf{85.01}} & \multirow{3}{*}{\textbf{0.50}}\\
\cline{1-7} \cline{1-7}
FLDCRF & 2/1 & 2/2 & 2/3 & 2/4 & 3/1 & 3/2 & \multicolumn{1}{c|}{} & \multicolumn{1}{c|}{} & \multicolumn{1}{c|}{}\\ 
\cline{1-7}
F1 & 86.44 & 86.44 & 86.5 & \textcolor{gray}{\textbf{85.01}} & 86.6 & 85.89 & \multicolumn{1}{c|}{} & \multicolumn{1}{c|}{} & \multicolumn{1}{c|}{}\\
\hline \hline \hline
\multicolumn{10}{|c|}{Set 7} \\
\hline \hline
FLDCRF & 1/1 & 1/2 & 1/3 & 1/4 & 1/5 & 1/6 & Best & Worst & Std\\ 
\hline
F1 & \textcolor{gray}{\textbf{88.3}} & 89.05 & 89.19 & 88.7 & 88.52 & 88.96 & \multirow{3}{*}{89.23} & \multirow{3}{*}{\textbf{88.3}} & \multirow{3}{*}{\textbf{0.28}}\\
\cline{1-7} \cline{1-7}
FLDCRF & 2/1 & 2/2 & 2/3 & 2/4 & 3/1 & 3/2 & \multicolumn{1}{c|}{} & \multicolumn{1}{c|}{} & \multicolumn{1}{c|}{}\\ 
\cline{1-7}
F1 & 89.05 & 88.94 & \textbf{89.23} & 89.11 & 88.94 & 89.09  & \multicolumn{1}{c|}{} & \multicolumn{1}{c|}{} & \multicolumn{1}{c|}{}\\
\hline
\end{tabular}
\end{center}
\end{table}

\begin{table}[h!]
\caption{LSTM validation performance on nested validation sets 4 and 7 of the nested CV experiment on UCI gesture phase data. $N_{els}$ denotes number of epochs trained and $N_{hls}$ represents the number of hidden units in the model.}
\label{tab:LSTM-validation-nested-gesture}
\begin{center}
\begin{tabular}{| >{\centering\arraybackslash} m{1.45cm}| >{\centering\arraybackslash} m{0.62cm}| >{\centering\arraybackslash} m{0.62cm}| >{\centering\arraybackslash} m{0.62cm}| >{\centering\arraybackslash} m{0.62cm}| >{\centering\arraybackslash} m{0.62cm}| >{\centering\arraybackslash} m{0.62cm}| >{\centering\arraybackslash} m{0.62cm}| >{\centering\arraybackslash} m{0.62cm}| >{\centering\arraybackslash} m{0.62cm}| >{\centering\arraybackslash} m{0.9cm}| >{\centering\arraybackslash} m{0.8cm}| >{\centering\arraybackslash} m{0.8cm}|} 
\hline
\multicolumn{13}{|c|}{Set 4} \\
\hline  \hline
\backslashbox{$N_{els}$\kern-1em}{\kern-1em$N_{hls}$} & 2 & 5 & 7 & 10 & 14 & 20 & 50 & 100 & 200 & Best & Worst & Std\\ 
\hline \hline
100 & 86.3 & 86.7 & 86.3 & 87.2 & 85.6 & 83.4 & 84.6 & 85.3 & 84.6 & \multirow{5}{*}{\textbf{87.96}} & \multirow{5}{*}{81.64}  & \multirow{5}{*}{1.43}\\
\cline{1-10}
200 & 85.3 & 87.3 & 86.3 & 84.6 & 85.6 & 86.6 & 82.5 & 86.0 & 83.9 & \multicolumn{1}{c|}{} & \multicolumn{1}{c|}{} & \multicolumn{1}{c|}{}\\ 
\cline{1-10}
300 & 85.0 & \textbf{87.8} & 85.5 & 87.9 & 85.3 & 87.5 & \textbf{85.5} & 86.6 & 85.5 & \multicolumn{1}{c|}{} & \multicolumn{1}{c|}{} & \multicolumn{1}{c|}{}\\
\cline{1-10}
400 & 84.8 & \textcolor{gray}{\textbf{85.7}} & 86.2 & 86.7 & 85.9 & \textbf{87.9} & \textcolor{gray}{\textbf{83.5}} & \textbf{86.3} & 85.7 & \multicolumn{1}{c|}{} & \multicolumn{1}{c|}{} & \multicolumn{1}{c|}{}\\
\cline{1-10}
500 & 85.5 & 	86.0 & 84.7 & 86.1 & 85.3 & \textcolor{gray}{\textbf{83.1}} & 83.5 & \textcolor{gray}{\textbf{81.6}} & 86.4 & \multicolumn{1}{c|}{} & \multicolumn{1}{c|}{} & \multicolumn{1}{c|}{}\\
\hline \hline
Std & 0.59 & 0.71 & 0.69 & 1.23 & 0.28 & 2.31 & 1.14 & 2.05 & 0.99  &  &  & \multicolumn{1}{c|}{}\\
\hline \hline \hline
\multicolumn{13}{|c|}{Set 7} \\
\hline  \hline
\backslashbox{$N_{els}$\kern-1em}{\kern-1em$N_{hls}$} & 2 & 5 & 7 & 10 & 14 & 20 & 50 & 100 & 200 & Best & Worst  & Std\\ 
\hline \hline
100 & 86.6 & 86.9 & 86.2 & 86.4 & 86.6 & 86.8 & 87.7 & \textbf{90.1} & 88.6 & \multirow{5}{*}{\textbf{90.72}} & \multirow{5}{*}{80.31} & \multirow{5}{*}{1.65} \\
\cline{1-10}
200 & 86.5 & 88.5 & 86.6 & 85.7 & 86.7 & 89.3 & 86.7 & 88.2 & 87.7 & \multicolumn{1}{c|}{} & \multicolumn{1}{c|}{} & \multicolumn{1}{c|}{}\\ 
\cline{1-10}
300 & 88.3 & 89.0 & 87.2 & 86.8 & 85.9 & 90.1 & 86.2 & 88.3 & 87.8 & \multicolumn{1}{c|}{} & \multicolumn{1}{c|}{} & \multicolumn{1}{c|}{}\\
\cline{1-10}
400 & 87.3 & 87.9 & 87.5 & 86.5 & 86.9 & \textbf{90.7} & 86.9 & 88.1 & 87.4  & \multicolumn{1}{c|}{} & \multicolumn{1}{c|}{} & \multicolumn{1}{c|}{}\\
\cline{1-10}
500 & 85.7 & 88.7 & 88.2 & 87.2 & 87.6 & \textcolor{gray}{\textbf{80.3}} & 87.0 & \textcolor{gray}{\textbf{87.1}} & 88.2 & \multicolumn{1}{c|}{} & \multicolumn{1}{c|}{} & \multicolumn{1}{c|}{}\\
\hline \hline
Std & 0.99 & 0.83 & 0.78 & 0.56 & 0.59 & 4.27 & 0.55 & 1.08 & 0.47  &  &  & \multicolumn{1}{c|}{}\\
\hline
\end{tabular}
\end{center}
\end{table}

Table \ref{tab:nested-cv-best-inner-gesture} highlights the overall FLDCRF-s and LSTM cross-validation performance on the inner loops (for selecting models). We report the best, worst case performances and standard deviation among models on validation sets. We also report percentage of poor performing models, i.e., validated models (hyperparameters) below a given threshold (F1-score 85\%). FLDCRF-s produces the lower worst case performance with lesser standard deviation among models than LSTM in most of the sets. 

High percentage of LSTM models performing below threshold across most of the sets is also noticeable and brings about difficulty to select the optimum models, by requiring careful monitoring and having to consider widespread hyperparameter settings. 5\% of all considered FLDCRF-s models across 7 cross-validation sets report below F1-score 85, as compared to 17.1\% for LSTM models. In set 7, although 2.2\% of LSTM models report F1 below 85, a staggering 69\% of the models perform below F1-score 88, while FLDCRF-s has a worst case F1 of 88.3.

\begin{table}[h!]
\caption{Summary of cross-validation performance by FLDCRF-s and LSTM on the inner loops of nested CV experiment on the UCI gesture phase data.}
\label{tab:nested-cv-best-inner-gesture}
\begin{center}
\begin{tabular}{| >{\centering\arraybackslash} m{3.5cm}| >{\centering\arraybackslash} m{1cm}| >{\centering\arraybackslash} m{1cm} | >{\centering\arraybackslash} m{1cm}| >{\centering\arraybackslash} m{1cm}| >{\centering\arraybackslash} m{1cm}| >{\centering\arraybackslash} m{1cm}| >{\centering\arraybackslash} m{1cm}|} 
\hline
\multirow{2}{*}{Model} & \multicolumn{7}{c|}{F1-score (best)} \\ [0.4ex] 
\cline{2-8}
\multicolumn{1}{|c|}{} & 1 & 2 & 3 & 4 & 5 & 6 & 7 \\
\hline
FLDCRF-s & 87.44 & 86.55 & 88.88 & 86.73 & \textbf{88.59} & 86.32 & 89.23 \\
\hline
LSTM & \textbf{88.24} & \textbf{87.32} & \textbf{89.85} & \textbf{87.96} & 86.71 & \textbf{87.27} & \textbf{90.72} \\
\hline \hline
\multirow{2}{*}{Model} & \multicolumn{7}{c|}{F1-score (worst)} \\ [0.4ex] 
\cline{2-8}
\multicolumn{1}{|c|}{} & 1 & 2 & 3 & 4 & 5 & 6 & 7 \\
\hline
FLDCRF-s & \textbf{86.02}    &    \textbf{84.38}    &    84.89    &   \textbf{85.01}     &   \textbf{86.91}   &     \textbf{85.03}    &     \textbf{88.3} \\
\hline
LSTM & 84.24    &    82.17     &   \textbf{85.17}    &    81.64   &    82.24   &     78.58    &    80.31 \\
\hline \hline
\multirow{2}{*}{Model} & \multicolumn{7}{c|}{F1-score (std)} \\ [0.4ex] 
\cline{2-8}
\multicolumn{1}{|c|}{} & 1 & 2 & 3 & 4 & 5 & 6 & 7 \\
\hline
FLDCRF-s & \textbf{0.38} & 0.83 & 1.29 & \textbf{0.55} & \textbf{0.47} & \textbf{0.50} & \textbf{0.27}	 \\
\hline
LSTM & 0.90 & \textbf{0.70} & \textbf{1.07} & 1.43 & 0.94 & 1.36 & 1.65 \\
\hline \hline
\multirow{2}{*}{Model} & \multicolumn{7}{c|}{\# F1-score $<$ 85 (\% of all validated models)} \\ [0.4ex] 
\cline{2-8}
\multicolumn{1}{|c|}{} & 1 & 2 & 3 & 4 & 5 & 6 & 7 \\
\hline
FLDCRF-s & \textbf{0} & 28.5 & 7.14 & \textbf{0} & \textbf{0} & \textbf{0} & \textbf{0}	 \\
\hline
LSTM & 8.9 & \textbf{20} & \textbf{0} & 26.7 & 37.8 & 24.4 & 2.2 \footnotemark	 \\
\hline 
\end{tabular}
\end{center}
\end{table}

\footnotetext{68.9\% of the LSTM models below F1-score 88, while FLDCRF-s reports worst case F1 of 88.3. }

\item \textit{Consistency:}

LSTM outperforms FLDCRF-s on the best cross-validation performances (see F1-score (best) in Table \ref{tab:nested-cv-best-inner-gesture}) on most (6 out of 7) of the sets. However, such optimized models fail to consistently outperform FLDCRF-s on the test sets (see Table \ref{tab:nested-cv-test-gesture}), and manage to perform better only on 3 (out of 7) sets. The average performance reported by LSTM on the test data is also about 0.5\% less than FLDCRF-s. 

\item \textit{Computation times:} 

Figure \ref{training-time-nested-gesture} compares considered FLDCRF-s and LSTM training times. LSTM training times are shown for 500 training epochs, while the FLDCRF-s training times are until convergence (by the default criteria for BFGS in Stan). Displayed results are quite similar to experiment 1, only magnified by the amount of training data. As mentioned earlier, FLDCRF training times can be further reduced by GPU implementation.
\end{itemize}

\begin{figure}[h]
\includegraphics[scale=0.7]{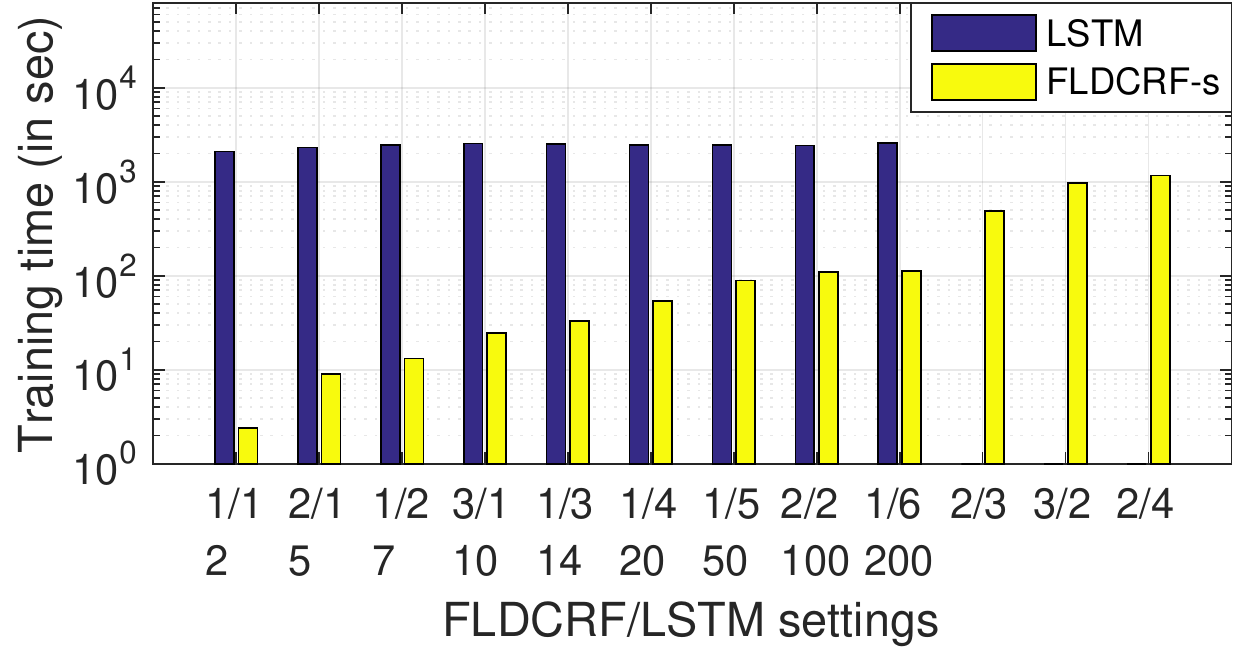}
\centering
\caption{Average training time required by different FLDCRF settings per outer fold of the nested CV experiment on the UCI gesture phase data. }
\label{training-time-nested-gesture}
\end{figure}

\subsection{UCI Opportunity Dataset}  \label{results-opportunity}

We perform our multi-label sequence tagging experiment on the UCI Opportunity Dataset. We compare several single-label (CRF, LDCRF, FLDCRF-s, LSTM, LSTM-CRF) and multi-label (FCRF, CCRF, FLDCRF-m1, FLDCRF-m2, LSTM-m) models in the nested CV experiment. 

\subsubsection{Experiment 1: nested CV}

This experiment has 5 outer loops, one for testing on each ADL sequence (see Section \ref{subsec:opportunity-data}). In each case, we select the models by 4-fold cross-validation on the remaining 4 ADL sequences. Sequence details for this experiment are illustrated in Table \ref{tab:UCI-oppor-sequence-details}. Tabulated data (features and labels) for this experiment is available here\footnote{https://github.com/satyajitneogiju/FLDCRF-for-sequence-labeling}.

\begin{itemize}[leftmargin=*]
\item \textit{Test performance:} 

We present individual labeling results (F1-scores) from different models in Tables \ref{tab:oppor-loco-outer} (for locomotion activity) and \ref{tab:oppor-hl-outer} (for HL activity). The results for the joint labeling task are presented in Table \ref{tab:oppor-overall-outer}.

Similar to the nested CV experiment on the UCI gesture phase data, FLDCRF-s improves the overall LDCRF performance on the dataset and outperforms LSTM and all other models. LSTM with the regular softmax layer for classification outperforms other LSTM models and LDCRF. LDCRF does not significantly improve CRF performance by modeling the latent dynamics, neither do the multi-label models FCRF, CCRF by considering joint learning. However, FLDCRF-m2 achieves notable improvement ($\sim$0.4\% on average) over these models by considering latent-dynamic interactions (cotemporal and first-order Markov, see Section \ref{subsec:Models}) among the label categories (locomotion and HL). It should be noted that, multiple hidden layers (1-3) for the same activity label were involved in the validation process of FLDCRF-s. So, multiple hidden layers for each label category can be employed in a FLDCRF-m2 model in order to further improve performance. FLDCRF-m1 only includes co-temporal interaction among the hidden layers (see Section \ref{subsec:Models}), and achieves marginal improvement over other CRF models.

\begin{table}[h]
\caption{Different model performances on the `locomotion' activity label during nested CV experiment on UCI opportunity data.}
\label{tab:oppor-loco-outer}
\begin{center}
\begin{tabular}{| >{\centering\arraybackslash} m{3.5cm}| >{\centering\arraybackslash} m{1cm}| >{\centering\arraybackslash} m{1cm} | >{\centering\arraybackslash} m{1cm}| >{\centering\arraybackslash} m{1cm}| >{\centering\arraybackslash} m{1cm}| >{\centering\arraybackslash} m{1.5cm}|} 
\hline
\multirow{2}{*}{Model} & \multicolumn{6}{c|}{F1 score (Locomotion)} \\ [0.4ex] 
\cline{2-7}
\multicolumn{1}{|c|}{} & 1 & 2 & 3 & 4 & 5 & Average \\
\hline
CRF & 48.26 & 64.4 & 84.01 & 82.32 & 86.87 & 73.17 \\
\hline
LDCRF & \textbf{48.3} & 64.4 & 84.04 & 82.32 & 86.87 & 73.19 \\
\hline
FLDCRF-s & \textbf{48.3} & 64.4 & 84.04 & \textbf{85.83} & 86.87 & \textbf{73.89} \\
\hline
FCRF & 48.16 & 64.42 & 83.93 & 82.53 & 86.84 & 73.18 \\
\hline
CCRF & 48.16 & 64.32 & 83.95 & 82.56 & 86.74 & 73.15 \\
\hline
FLDCRF-m1 & 48.23 & 64.4 & 83.98 & 82.84 & 86.87 & 73.26 \\
\hline
FLDCRF-m2 & 48.23 & \textbf{64.52} & \textbf{84.23} & 83.88 & \textbf{86.93} & 73.56 \\
\hline \hline
LSTM & 46.41 & \textbf{64.28} & 77.64 & \textbf{91.73} & 87.94 & \textbf{73.60} \\
\hline
LSTM-CRF &  46.02 & 59.98 & 73.48 & 89.19 & 81.16 & 69.97   \\
\hline
LSTM-m &  \textbf{46.60} & 61.47 & \textbf{78.66} & 90.63 & \textbf{88.38}	& 73.15   \\
\hline
\end{tabular}
\end{center}
\end{table}

Table \ref{tab:oppor-hl-outer} presents the labeling performance of different models on continuous tagging of the high-level (HL) activity. LDCRF, FLDCRF-s, FCRF and CCRF produce similar overall performance, marginally improving over the simple LCCRF (or CRF) model. LSTM-CRF outperforms LSTM and LSTM-m models and achieves a similar overall performance to that of the LCCRF model. FLDCRF-m2, by considering the latent-dynamic interactions in the joint labeling task, significantly (by atleast 0.75\% on average) outperforms all CRF and LSTM models, with significant ($>$3\%) improvement over LDCRF on set 1. As in case of locomotion, FLDCRF-m1 achieves marginal improvement over other CRF models.

\begin{table}[h]
\caption{Different model performances on the `high-level' activity (HL) label during nested CV experiment on UCI opportunity data.}
\label{tab:oppor-hl-outer}
\begin{center}
\begin{tabular}{| >{\centering\arraybackslash} m{3.5cm}| >{\centering\arraybackslash} m{1cm}| >{\centering\arraybackslash} m{1cm} | >{\centering\arraybackslash} m{1cm}| >{\centering\arraybackslash} m{1cm}| >{\centering\arraybackslash} m{1cm}| >{\centering\arraybackslash} m{1.5cm}|} 
\hline
\multirow{2}{*}{Model} & \multicolumn{6}{c|}{F1 score (HL)} \\ [0.4ex] 
\cline{2-7}
\multicolumn{1}{|c|}{} & 1 & 2 & 3 & 4 & 5 & Average \\
\hline
CRF & 80.29 & 97.43 & 99.57 & \textbf{100} & \textbf{100} & 95.46 \\
\hline
LDCRF & 81.04 & 97.57 & 99.57 & \textbf{100} & \textbf{100} & 95.64 \\
\hline
FLDCRF-s & 81.04 & 97.55 & 99.56 & \textbf{100} & \textbf{100} & 95.63 \\
\hline
FCRF & 80.92 & 97.7 & 99.6 & \textbf{100} & \textbf{100} & 95.64 \\
\hline
CCRF & 80.91& 97.71 & 99.57 & \textbf{100} & \textbf{100} & 95.64 \\
\hline
FLDCRF-m1 & 81.38 & 97.73 & \textbf{99.72} & 99.88 & \textbf{100} & 95.74 \\
\hline
FLDCRF-m2 & \textbf{84.29} & \textbf{98.1} & 99.56 & \textbf{100} & \textbf{100} & \textbf{96.39} \\
\hline  \hline
LSTM & \textbf{94.26} & 90.57 & \textbf{100} & \textbf{100} & 89.04 & 94.77 \\
\hline
LSTM-CRF &  81.21 & \textbf{98.04} & \textbf{100} & \textbf{100} & 98.02 & \textbf{95.45}  \\
\hline
LSTM-m & 80.05 & 97.01 & 99.57 & \textbf{100} & \textbf{98.28} & 94.98 \\
\hline
\end{tabular}
\end{center}
\end{table}

Table \ref{tab:oppor-overall-outer} summarizes the joint labeling performance of different models. FLDCRF-s performs best among the single-label sequence models, while FLDCRF-m2 outperforms all single and multi-label sequence models. LSTM models fail to perform consistently across the test sets, and produces realtively poorer overall performance. The multi-label LSTM-m achieves best results among the LSTM models. FLDCRF-m2 improves the LDCRF, FCRF, CCRF and CRF performance on all sets, demonstrating the significance of modeling latent-dynamic interactions among different label categories. FLDCRF-m2 outshines the LSTM model on 3 (out of 5) sets, and outperforms LSTM-CRF and LSTM-m models on 4 sets each, with significantly outperforming (by $>$1.5\%) all the models on the entire dataset. 

\begin{table}[h]
\caption{Overall model performances on the joint sequence labeling task during nested CV experiment on UCI opportunity data.}
\label{tab:oppor-overall-outer}
\begin{center}
\begin{tabular}{| >{\centering\arraybackslash} m{3.5cm}| >{\centering\arraybackslash} m{1cm}| >{\centering\arraybackslash} m{1cm} | >{\centering\arraybackslash} m{1cm}| >{\centering\arraybackslash} m{1cm}| >{\centering\arraybackslash} m{1cm}| >{\centering\arraybackslash} m{1.5cm}|} 
\hline
\multirow{2}{*}{Model} & \multicolumn{6}{c|}{F1 score (Overall)} \\ [0.4ex] 
\cline{2-7}
\multicolumn{1}{|c|}{} & 1 & 2 & 3 & 4 & 5 & Average \\
\hline
CRF & 57.67 & 79.69 & 91.58 & 91.16 & 93.43 & 82.71 \\
\hline
LDCRF & 58.21 & 79.83 & 91.59 & 91.16 & 93.43 & 82.84 \\
\hline
FLDCRF-s & 58.21 & 79.80 & 91.58 & \textbf{92.92} & 93.45 & 83.19 \\
\hline
FCRF & 58.05 & 79.97 & 91.56 & 91.27 & 93.42 & 82.85 \\
\hline
CCRF & 58.05 & 79.93 & 91.55 & 91.28 & 93.37 & 82.84 \\
\hline
FLDCRF-m1 & 58.42 & 79.98 & \textbf{91.71} & 91.3 & 93.43 & 82.97 \\
\hline
FLDCRF-m2 & \textbf{60.53} & \textbf{80.4} & 91.67 & 91.94 & \textbf{93.47} & \textbf{83.60} \\
\hline  \hline
LSTM & \textbf{67.78} & 73.52 & 88.82 & \textbf{95.86} & 84.09 & 82.01 \\
\hline
LSTM-CRF &  57.19 & \textbf{78.07} & 86.74 & 94.60 & 88.64 & 81.05 \\
\hline
LSTM-m &  56.67 & 77.83 & \textbf{88.90} & 95.31 & \textbf{92.50} & \textbf{82.24}  \\
\hline
\end{tabular}
\end{center}
\end{table}

\item \textit{Model selection:} 

We present the validation performance of the best models from each family, FLDCRF-m (written for FLDCRF-m2) and LSTM-m, on cross-validation sets 2 and 5 in Tables \ref{oppor-validation-set1-FLDCRF} and \ref{oppor-validation-set1-LSTM} respectively. We omit the detailed results for other sets to save space, and summarize all the best, worst and standard deviation performances on the inner loops of nested CV in Table \ref{oppor-validation}.

LSTM-m gives rapidly fluctuating validation performance across epochs for most of the considered $N_{hls}$ setups (see Table \ref{oppor-validation-set1-LSTM}), especially during cross-validation on set 5. It also produces significantly lower worst case performance with high percentage of models giving notably poor results (19.5\% of all models across 5 sets with F1$<$72, see Table \ref{oppor-validation} for individual sets). FLDCRF-m, on the other hand, shows much discipline against the choices of the hyperparameters, and shows decline in performance beyond certain $N_h/N_s$ settings (see Tables \ref{oppor-validation-set1-FLDCRF} and \ref{oppor-validation}), making it much easier to work with.

\begin{table}[h!]
\caption{FLDCRF-m2 validation performance on nested cross-validation sets 2 and 5 on UCI opportunity data.}
\label{oppor-validation-set1-FLDCRF}
\begin{center}
\begin{tabular}{| c| c| c| c| c| c| c| c| c| c|} 
\hline
\multicolumn{10}{|c|}{Set 2} \\
\hline \hline
FLDCRF & 1/1 & 1/2 & 1/3 & 1/4 & 1/5 & 1/6 & Best & Worst & Std\\ 
\hline
F1 & 76.17  &  76.18  &  76.23  &  76.2  &  76.63  &  76.2  & \multirow{5}{*}{76.84} & \multirow{5}{*}{\textbf{75.34}} & \multirow{5}{*}{\textbf{0.33}}\\
\cline{1-7} \cline{1-7}
FLDCRF & 2/1 & 2/2 & 2/3 & 2/4 & 2/5 & 2/6 & \multicolumn{1}{c|}{} & \multicolumn{1}{c|}{} & \multicolumn{1}{c|}{}\\ 
\cline{1-7}
F1 & 76.32  &  76.84  &  76.33  &  76.51  &  76.32  &  76.32 & \multicolumn{1}{c|}{} & \multicolumn{1}{c|}{} & \multicolumn{1}{c|}{}\\
\hline
FLDCRF & 3/1 & 3/2 & 3/3 &  &  &  & \multicolumn{1}{c|}{} & \multicolumn{1}{c|}{} & \multicolumn{1}{c|}{}\\ 
\cline{1-7}
F1 & 76.31  &  76.65  &  75.34  &    &    &    & \multicolumn{1}{c|}{} & \multicolumn{1}{c|}{} & \multicolumn{1}{c|}{}\\
\hline \hline \hline
\multicolumn{10}{|c|}{Set 5} \\
\hline \hline
FLDCRF & 1/1 & 1/2 & 1/3 & 1/4 & 1/5 & 1/6 & Best & Worst  & Std\\ 
\hline
F1 & 73.27  &  74.48  &  73.29  &  73.27  &  74.37  &  73.62  & \multirow{5}{*}{74.48} & \multirow{5}{*}{\textbf{72.78}} & \multirow{5}{*}{\textbf{0.46}}\\
\cline{1-7} \cline{1-7}
FLDCRF & 2/1 & 2/2 & 2/3 & 2/4 & 2/5 & 2/6  & \multicolumn{1}{c|}{} & \multicolumn{1}{c|}{} & \multicolumn{1}{c|}{}\\ 
\cline{1-7}
F1 & 73.32  &  73.36  &  74.11  &  73.3  &   73.3   &  73.53   & \multicolumn{1}{c|}{} & \multicolumn{1}{c|}{} & \multicolumn{1}{c|}{}\\
\hline
FLDCRF & 3/1 & 3/2 & 3/3 &  &  &   & \multicolumn{1}{c|}{} & \multicolumn{1}{c|}{} & \multicolumn{1}{c|}{}\\ 
\cline{1-7}
F1 &  73.33   & 73.31  &  72.78  &    &    &    & \multicolumn{1}{c|}{} & \multicolumn{1}{c|}{} & \multicolumn{1}{c|}{}\\
\hline
\end{tabular}
\end{center}
\end{table}

\begin{table}[h!]
\caption{LSTM-m validation performance on nested cross-validation sets 2 and 5 on UCI opportunity data.}
\label{oppor-validation-set1-LSTM}
\begin{center}
\begin{tabular}{| >{\centering\arraybackslash} m{1.45cm}| >{\centering\arraybackslash} m{0.62cm}| >{\centering\arraybackslash} m{0.62cm}| >{\centering\arraybackslash} m{0.62cm}| >{\centering\arraybackslash} m{0.62cm}| >{\centering\arraybackslash} m{0.62cm}| >{\centering\arraybackslash} m{0.62cm}| >{\centering\arraybackslash} m{0.62cm}| >{\centering\arraybackslash} m{0.62cm}| >{\centering\arraybackslash} m{0.9cm}| >{\centering\arraybackslash} m{0.8cm}| >{\centering\arraybackslash} m{0.8cm}|} 
\hline
\multicolumn{12}{|c|}{Set 2} \\
\hline \hline
\backslashbox{$N_{els}$\kern-1em}{\kern-1em$N_{hls}$} & 5 & 10 & 25 & 50 & 75 & 150 & 300 & 500  & Best & Worst & Std\\ 
\hline \hline
100 & \textbf{81.1}  &   79.4   &  79.6  &   79.7  &   79.3   &  \textbf{81.1}  &   81.3   &   70.1  & \multirow{5}{*}{\textbf{83.23}} & \multirow{5}{*}{67.98} & \multirow{5}{*}{3.49} \\
\cline{1-9}
200 & \textcolor{gray}{\textbf{74.0}}  &   80.2   &  83.2  &   76.4   &  \textcolor{gray}{\textbf{75.6}}   &  \textcolor{gray}{\textbf{78.2}}  &   75.0  &   73.6  & \multicolumn{1}{c|}{} & \multicolumn{1}{c|}{} & \multicolumn{1}{c|}{}\\ 
\cline{1-9}
300 & \textbf{80.5}  &   77.6  &   81.8   &  75.5   &  \textbf{80.3}  &   78.2  &   80.8   &  72.5 & \multicolumn{1}{c|}{} & \multicolumn{1}{c|}{} & \multicolumn{1}{c|}{}\\
\cline{1-9}
400 & \textcolor{gray}{\textbf{75.3}}   &   75.1  &   77.5  &   74.3   &  \textcolor{gray}{\textbf{77.6}}   &  \textcolor{gray}{\textbf{76.1}}   &  76.5   &  68.1 & \multicolumn{1}{c|}{} & \multicolumn{1}{c|}{} & \multicolumn{1}{c|}{}\\
\cline{1-9}
500 & 77.6  &   75.9  &   77.2  &   76.2  &   \textcolor{gray}{\textbf{74.7}}  &   76.1  &   75.9   &  67.9 & \multicolumn{1}{c|}{} & \multicolumn{1}{c|}{} & \multicolumn{1}{c|}{}\\
\hline \hline \hline
\multicolumn{12}{|c|}{Set 5} \\
\hline \hline
\backslashbox{$N_{els}$\kern-1em}{\kern-1em$N_{hls}$} & 5 & 10 & 25 & 50 & 75 & 150 & 300 & 500  & Best & Worst & Std\\ 
\hline \hline
100 & 71.6   &   \textbf{73.3}   &  71.7  &   77.4  &   68.6  &   \textbf{81.1}  &   72.2  &   77.6  & \multirow{5}{*}{\textbf{85.7}} & \multirow{5}{*}{67.49} & \multirow{5}{*}{4.17} \\
\cline{1-9}
200 & 75.0 &    \textcolor{gray}{\textbf{70.9}}  &   71.3  &   \textbf{80.6}  &   \textcolor{gray}{\textbf{67.4}}  &   \textcolor{gray}{\textbf{73.6}}  &   72.2   &  78.9   & \multicolumn{1}{c|}{} & \multicolumn{1}{c|}{} & \multicolumn{1}{c|}{}\\ 
\cline{1-9}
300 & \textbf{78.3}   &  71.1   &  \textcolor{gray}{\textbf{70.1}}  &   \textcolor{gray}{\textbf{71.4}}   &  70.0  &   70.1  &   69.2  &   79.6  & \multicolumn{1}{c|}{} & \multicolumn{1}{c|}{} & \multicolumn{1}{c|}{}\\
\cline{1-9}
400 & \textcolor{gray}{\textbf{71.6}}  &   70.4  &   \textbf{78.8} &   72.0   &  \textcolor{gray}{\textbf{71.3}}    &    \textcolor{gray}{\textbf{74}}   &   72.2  &   78.9  & \multicolumn{1}{c|}{} & \multicolumn{1}{c|}{} & \multicolumn{1}{c|}{}\\
\cline{1-9}
500 & 72.1  &   71.4  &   \textcolor{gray}{\textbf{73.2}}   &  72.3   &   \textbf{78.8}   &   \textbf{85.7}  &   70.8  &    79.6 & \multicolumn{1}{c|}{} & \multicolumn{1}{c|}{} & \multicolumn{1}{c|}{}\\
\hline
\end{tabular}
\end{center}
\end{table}

\begin{table}[h!]
\caption{Summary of cross-validation performance by FLDCRF-s and LSTM on the inner loops of nested CV experiment on the UCI opportunity data.}
\label{oppor-validation}
\begin{center}
\begin{tabular}{| >{\centering\arraybackslash} m{3.5cm}| >{\centering\arraybackslash} m{1cm}| >{\centering\arraybackslash} m{1cm} | >{\centering\arraybackslash} m{1cm}| >{\centering\arraybackslash} m{1cm}| >{\centering\arraybackslash} m{1cm}|} 
\hline
\multirow{2}{*}{Model} & \multicolumn{5}{c|}{F1 score (best)} \\ [0.4ex] 
\cline{2-6}
\multicolumn{1}{|c|}{} & 1 & 2 & 3 & 4 & 5 \\
\hline
FLDCRF-m2 & 87.54 & 76.84 & 75.12 & 74.31 & 74.48	 \\
\hline
LSTM-m & \textbf{87.70} &  \textbf{83.23} & \textbf{79.45} & \textbf{80.15}  & \textbf{85.70}	 \\
\hline \hline
\multirow{2}{*}{Model} & \multicolumn{5}{c|}{F1 score (worst)} \\ [0.4ex] 
\cline{2-6}
\multicolumn{1}{|c|}{} & 1 & 2 & 3 & 4 & 5 \\
\hline
FLDCRF-m2 & 86.96 & 75.34  &   73.25 &   72.67 &   72.78	 \\
\hline
LSTM-m &  \textcolor{gray}{\textbf{79.26}} &   \textcolor{gray}{\textbf{67.98}} &  \textcolor{gray}{\textbf{68.81}} &  \textcolor{gray}{\textbf{65.23}} &   \textcolor{gray}{\textbf{67.49}}	 \\
\hline \hline
\multirow{2}{*}{Model} & \multicolumn{5}{c|}{F1 score (std)} \\ [0.4ex] 
\cline{2-6}
\multicolumn{1}{|c|}{} & 1 & 2 & 3 & 4 & 5 \\
\hline
FLDCRF-m2 & \textbf{0.17} & \textbf{0.33} & \textbf{0.66} & \textbf{0.45} & \textbf{0.46} \\
\hline
LSTM-m & 2.00 & 3.49 & 3.35 & 4.26 & 4.17 \\
\hline
\hline \hline
\multirow{2}{*}{Model} & \multicolumn{5}{c|}{\% F1 $<$ 72} \\ [0.4ex] 
\cline{2-6}
\multicolumn{1}{|c|}{} & 1 & 2 & 3 & 4 & 5 \\
\hline
FLDCRF-m2 & 0 & 0 & 0 & 0 & 0 \\
\hline
LSTM-m & 0 & \textcolor{gray}{\textbf{10}} & \textcolor{gray}{\textbf{17.5}} & \textcolor{gray}{\textbf{27.5}} & \textcolor{gray}{\textbf{42.5}} \\
\hline
\end{tabular}
\end{center}
\end{table}

\item \textit{Consistency:}

As in our earlier experiments, LSTM models (LSTM-m) are considerably better than FLDCRF-m to give the best validation results across all sets (see Table \ref{oppor-validation}). However, such optimized LSTM-m models fail to consistently beat FLDCRF-m on the test sets, rather producing inconsistent performance across test sets (managing superior test performance only on 1 out of 5 sets, see Table \ref{tab:oppor-overall-outer}). This is true for all other LSTM variants tested in this paper. Such inconsistency among validation and test performance raises serious concerns about practical applications of LSTM models.

\item \textit{Computation times:} 

Figure \ref{fig:FLDCRF-time-oppor} compares training and inference (to include) times required by different FLDCRF-m and LSTM-m models. As mentioned earlier, FLDCRF-m is run on a CPU and can be made faster by GPU implementation.


\begin{figure}[h!]
\includegraphics[scale=0.6]{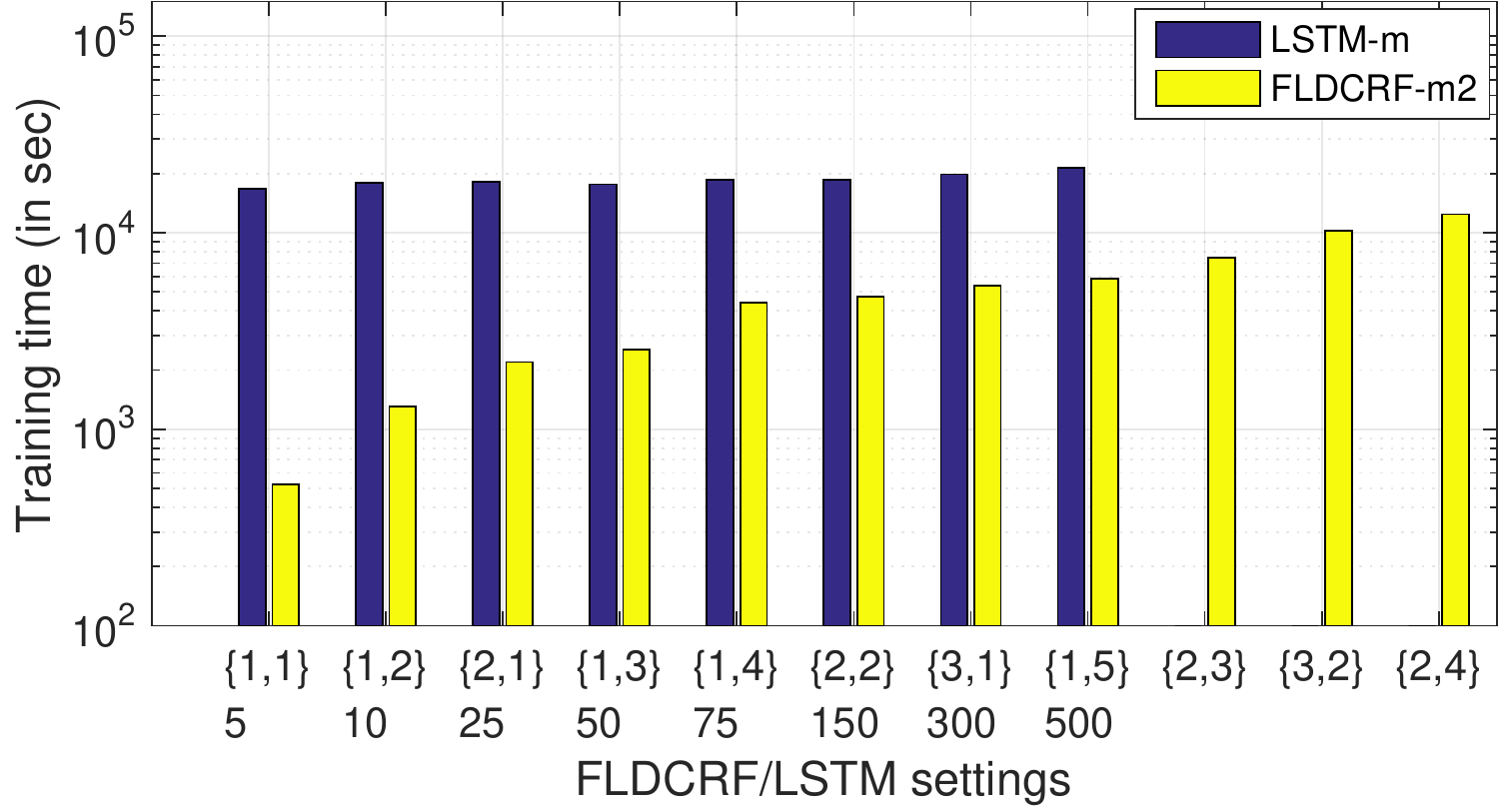}
\centering
\caption{Average training time required by FLDCRF-m and LSTM-m per outer fold of UCI Opportunity data.}
\label{fig:FLDCRF-time-oppor}
\end{figure}

\end{itemize}

\subsection{Multi-view Experiment}  \label{subsec:multi-view}

In this section, we examine the multi-view LDCRFs \citep{Coupled-LDCRF} (see Fig. \ref{fig:CLDCRF}) on the UCI opportunity data. We perform nested CV experiments separately on the locomotion and HL activity labels of the UCI opportunity data. As described in Section \ref{subsec:opportunity-data}, the feature set $x_t$ for opportunity data comprises of 113 body-worn sensor features ($x_{1,t}$) and 32 shoe sensor ($x_{2,t}$) features i.e., $x_t = \{x_{1,t}, x_{2,t}\}$. We consider three different LDCRFs (with observations $x_{1,t}$, $x_{2,t}$ and $x_t$ respectively, see Fig. \ref{fig:CDCRF}c) and a coupled-linked MVLDCRF for comparison. We also consider the FLDCRF-s performance with $x_t$. We present results of compared models on 10 test sets in Table \ref{tab:multi-view}.

LDCRF-body, LDCRF-shoe and LDCRF-(body+shoe) correponds to LDCRFs with observations $x_{1,t}$, $x_{2,t}$ and $x_t$ respectively. LDCRF-averaged gives the mean of LDCRF-body and LDCRF-shoe performances. MVLDCRF-averaged performance correponds to the training and weighted inference mechanism by \citet{Coupled-LDCRF}, while MVLDCRF-body layer and MVLDCRF-shoe layer gives the F1 scores obtained from individual layers of Fig. \ref{fig:CLDCRF}.


\begin{table}[h!]
\caption{Comparing different models by distributing body-worn (113) and shoe sensor (32) features. LDCRF-body, LDCRF-shoe and LDCRF-(body+shoe) correponds to $x_{1,t}$, $x_{2,t}$ and $x_t$ respectively. LDCRF-averaged gives the mean of LDCRF-body and LDCRF-shoe outputs. MVLDCRF-averaged performance correponds to the training and weighted inference mechanism by \citet{Coupled-LDCRF}, while MVLDCRF-body layer and MVLDCRF-shoe layer gives the F1-scores obtained from individual layers of Fig. \ref{fig:CLDCRF}.}
\label{tab:multi-view}
\begin{center}
\begin{tabular}{| >{\centering\arraybackslash} m{3.5cm}| >{\centering\arraybackslash} m{1.3cm}| >{\centering\arraybackslash} m{1.3cm}|  >{\centering\arraybackslash} m{1.3cm}| >{\centering\arraybackslash} m{1.3cm}| >{\centering\arraybackslash} m{1.3cm}| >{\centering\arraybackslash} m{2cm}|} 
\hline
\multirow{2}{*}{Model} & \multicolumn{6}{c|}{HL} \\ [0.4ex] 
\cline{2-7}
\multicolumn{1}{|c|}{} & Set1 & Set2 & Set3 & Set4 & Set5 & Average\\
\hline \hline
LDCRF-body & 79.96 & 97.13 & 99.57 & 99.83 & 100 & 95.29 \\
\hline
LDCRF-shoe & \textbf{98.48} & 97.83  &  \textbf{99.81}  & 99.97 & 99.94 & \textbf{99.21} \\
\hline
LDCRF-averaged & 89.22    &    97.48     &   99.69    &     99.9   &     99.97 & 97.25 \\
\hline
LDCRF-(body+shoe) & 81.04 & 97.57 & 99.57 & \textbf{100} & \textbf{100} & 95.64 \\
\hline
FLDCRF-s(body+shoe) & 81.04 & 97.55 & 99.56 & \textbf{100} & \textbf{100} & 95.63 \\
\hline
MVLDCRF-body layer &  86.63    &    97.96    &    99.63     &     \textbf{100}    &      \textbf{100} & 96.84\\
\hline
MVLDCRF-shoe layer &  91.81    &    \textbf{98.18}    &    99.71      &    \textbf{100}     &     \textbf{100}  & 97.94\\
\hline
MVLDCRF-averaged & 89.22     &   \textbf{98.07}    &    99.67    &      \textbf{100}    &     \textbf{100} & 97.39\\
\hline \hline \hline
\multirow{2}{*}{Model} & \multicolumn{6}{c|}{Locomotion} \\ [0.4ex] 
\cline{2-7}
\multicolumn{1}{|c|}{} & Set1 & Set2 & Set3 & Set4 & Set5 & Average\\
\hline \hline
LDCRF-body & 46.6    &    60.33    &     81.5    &     \textbf{82.9}    &    \textbf{87.22}  & 71.71 \\
\hline
LDCRF-shoe & \textbf{51.55}   &     \textbf{71.97}    &    76.81    &    81.34   &   77.81  &  71.9 \\
\hline
LDCRF-averaged & 49.07    &    66.15   &    79.155   &     82.12    &   82.51 & 71.80 \\
\hline
LDCRF-(body+shoe) & 48.3 & 64.4 & \textbf{84.04} & 82.32 & 86.87 & 73.19 \\
\hline
FLDCRF-s(body+shoe) & 48.3 & 64.4 & \textbf{84.04} & \textbf{85.83} & 86.87 & \textbf{73.89} \\
\hline
MVLDCRF-body layer &  48.26    &    63.93    &    83.98     &   81.95    &    87.09 & 73.04\\
\hline
MVLDCRF-shoe layer &  45.92     &   63.58     &   83.98    &    82.05    &    86.99  & 72.50\\
\hline
MVLDCRF-averaged & 47.09   &    63.75   &     83.98     &      82    &    87.04 & 72.77\\
\hline
\end{tabular}
\end{center}
\end{table}

In most of the cases (sets 1, 3 of HL; set 5 of locomotion) where MVLDCRF-averaged \citep{Coupled-LDCRF} outperformed the LDCRF with $x_t$, i.e., LDCRF-(body+shoe), either LDCRF-body (set 5 of locomotion) or LDCRF-shoe (sets 1, 3 of HL) has outperformed all other models, including MVLDCRF-averaged. In most other cases, one of the LDCRFs (LDCRF-(body+shoe), LDCRF-body, LDCRF-shoe) has outperformed MVLDCRF-averaged. 

MVLDCRF-averaged marginally outperforms all 3 LDCRF models (i.e., with $x_{1,t}$, $x_{2,t}$ and $x_t$) only on set 2 of HL, where LDCRF-body and LDCRF-shoe performances are similar. We thus argue that an improvement by a MVLDCRF-averaged \citep{Coupled-LDCRF} over all three LDCRFs (with $x_{1,t}$, $x_{2,t}$ and $x_t$) is possible only when the two individual LDCRFs with $x_{1,t}$ and $x_{2,t}$ perform similarly, as in set 2 of HL. It is advisable to test the three LDCRFs (or FLDCRF-s for improved performance) in case a feature distribution ($x_{1,t}$ and $x_{2,t}$) is available, else utilize the entire feature set $x_t$ with a FLDCRF-s and let the model learn the different interacting latent dynamics within the features and labels. Figuring out the distributions ($x_{1,t}$, $x_{2,t}$ etc.) if any, where two LDCRFs perform similarly is very much the only way for a MVLDCRF to perform better than all three LDCRFs (and possibly FLDCRF-s). However such a task is quite tedious, if not impossible.


\section{Discussion}   \label{sec:Discussion}

We have shown difficulties to select LSTM models on validation data across our experiments in the paper. LSTM models, although producing some excellent performance on validation data for certain hyperparameter settings, fail to be consistent on the test data. FLDCRF outperforms LSTM on test data across all our experiments. Additionally, FLDCRF requires less computation times for training and inference, even without GPU implementation. Moreover, FLDCRF being a graphical model, it is very easy to include known dependency information among the variables for improved modeling. Such inclusion in LSTM is less straightforward. Motivation behind FLDCRF, its graph structure and the mathematical model all are lucidly defined and are derived from each other. Although, we agree on the motivation and the mathematical formulations of LSTM, the modeling process is not very lucid and hard to comprehend. We summarize several validated modeling attributes of FLDCRF and LSTM in Table \ref{tab:model-compare}, in terms of user convenience. 

With difficulties in model selection, inconsistency across validation and test data, longer training and average performance on test data, LSTM has concerns about its stability and reliability on practical deployment. On the other hand, FLDCRF models bring peace of mind with lucid intuition, ease of model selection, consistent and superior performance on test data with shorter training. 

FLDCRF subsumes major sequential CRF variants, viz., LCCRF, LDCRF and DCRF (see Section \ref{sec:Models}) and outperforms each of these state-of-the-art models across experiments in the paper. FLDCRF performance can be further improved by considering more generic variants (utilize past features $x_{t-1}$, $x_{t-2}$ etc. during modeling, second order Markov dependency along hidden layers, multiple hidden layers for each label category etc.). We look forward to GPU implementation of FLDCRF and apply on several sequence labeling tasks of computer vision and NLP.


\begin{table}[h!]
\caption{Comparing FLDCRF and LSTM modeling attributes. HP stands for hyperparameter. '++' stands for very convenient for user, '+' stands for convenient, '-' stands for inconvenient and '- -' stands for very inconvenient.}
\label{tab:model-compare}
\begin{center}
\begin{tabular}{| >{\arraybackslash} m{6.5cm}| >{\centering\arraybackslash} m{2cm}| >{\centering\arraybackslash} m{2cm}|} 
\hline
\multirow{2}{*}{Attribute} & \multicolumn{2}{c|}{Model} \\ [0.4ex] 
\cline{2-3}
\multicolumn{1}{|c|}{} & FLDCRF & LSTM \\ [0.4ex] 
\hline \hline
Test performance & + & - \\ [0.4ex] 
\hline
Ease of model selection & ++ & - - \\ [0.4ex] 
\hspace{2em}$\bullet$ Less types of HP & \hspace{1em}$\bullet$ + & \hspace{1em}$\bullet$ - \\ [0.4ex] 
\hspace{2em}$\bullet$ Rule to select HP & \hspace{0.6em}$\bullet$ - & \hspace{1em}$\bullet$ - \\ [0.4ex]
\hspace{2em}$\bullet$ Tune training epochs & \hspace{1em}$\bullet$ + & \hspace{1em}$\bullet$ - \\ [0.4ex]
\hspace{2em}$\bullet$ HP starting point & \hspace{1em}$\bullet$ + & \hspace{1em}$\bullet$ - \\ [0.4ex]
\hspace{2em}$\bullet$ HP ending point & \hspace{1em}$\bullet$ + & \hspace{1em}$\bullet$ - \\ [0.4ex]
\hspace{2em}$\bullet$ Pattern among HP & \hspace{1.7em}$\bullet$ ++ & \hspace{1.5em}$\bullet$ - - \\ [0.4ex]
\hspace{2em}$\bullet$ Worst validation case & \hspace{1em}$\bullet$ + & \hspace{1.5em}$\bullet$ - - \\ [0.4ex]
\hspace{2em}$\bullet$ \% of poor models & \hspace{1em}$\bullet$ + & \hspace{1em}$\bullet$ - \\ [0.4ex]
\hspace{2em}$\bullet$ Variance among models & \hspace{1em}$\bullet$ + & \hspace{1em}$\bullet$ - \\ [0.4ex]
\hspace{2em}$\bullet$ Best validation case & \hspace{0.6em}$\bullet$ - & \hspace{1.4em}$\bullet$ + \\ [0.4ex]
\hline
Consistency across validation and test & ++ & - - \\ [0.4ex] 
\hline
Computation time & ++ & - - \\ [0.4ex] 
\hline
Including known dependency & ++ & - - \\ [0.4ex] 
\hline
Intuitive & ++ & - - \\ [0.4ex] 
\hline
Practical deployment & ++ & - - \\ [0.4ex] 
\hline
\end{tabular}
\end{center}
\end{table}

\section{Conclusion}   \label{sec:conclude}

We proposed FLDCRF, a single and multi-label generalization of LDCRF. We presented 3 different FLDCRF variants for single-label (FLDCRF-s), multi-label (FLDCRF-m) and multi-agent (FLDCRF-i) sequence prediction/tagging tasks. FLDCRF-s allows multiple interacting latent dynamics of the class labels and extends the capability of LDCRF, as well as outperforms LSTM and LSTM-CRF across multiple datasets. FLDCRF-m introduces hidden variables in a DCRF to accommodate latent dynamic interactions among different label categories, thereby improving DCRF performance and outperforming all state-of-the-art sequence models including CRF, LDCRF, FCRF, LSTM, LSTM-m on a joint sequence labeling task. We also described the LSTM model selection difficulties and its inconsistent performance across validation and test data (summarized in Table \ref{tab:model-compare}). By contrast, FLDCRF presents easier model selection, provides consistency across validation and test data and gurantees good performance on random model selection. FLDCRF also offers lucid model intuition and user flexibility over approach. We look forward to GPU implementation of FLDCRF and compare FLDCRF and LSTM on larger sequence datasets. Another interesting topic for future research is to apply end-to-end models with FLDCRF succeeded  by CNN layers on popular computer vision problems like action recognition. We are also exploring the multi-agent FLDCRF-i model for joint intention prediction of pedestrians on crossing/not-crossing before Autonomous Vehicles. It is possible to extend the idea of factorized latent space in FLDCRF to heterogeneous (discrete and continuous) state space models. \\


\acks{We would like to thank Dr. Michael Hoy for his contributions in this work. This research is partially supported by the ST
Engineering-NTU Corporate Lab through the NRF corporate lab@university scheme.}


\newpage

\appendix

\section{Latent Dynamic Conditional Random Fields (LDCRF)} \label{Appen:LDCRF}

In this section, we will describe the LDCRF \citep{LDCRF} mathematical model.

\subsection{Model}

The task is to learn a probabilistic mapping between a time-series sequence of observed input features \textbf{x} = \{$x_1, x_2, ... , x_T$\} and a sequence of observed classifcation labels \textbf{y} = \{\(y_1, y_2, ... , y_T\)\}. \(y_t\) \(\in \) \(\Upsilon \), \(\forall j = 1,2, ... , T\), where \(\Upsilon \) is the set of classification labels. \textbf{h} = \{\(h_1, h_2, ... , h_T\)\} denotes the hidden layer for capturing intrinsic dynamics within each label. Each label $\ell$ \(\in \) \(\Upsilon \) is associated with a set of hidden states $\textit{$\mathcal{H}$}_{\ell}$. \textit{$\mathcal{H}$} is the set of all possible hidden states written as \textit{$\mathcal{H}$} = \(\bigcup_{\ell} \textit{$\mathcal{H}$}_{\ell} \). \(\textit{$\mathcal{H}$}_{\ell} \) are disjoint $\forall$$\ell$ $\in$ \(\Upsilon \). Each $h_t$ is restricted to belong to the set ${\mathcal{H}}_{y_t}$, i.e., \(h_t \) \(\in \) \(\textit{$\mathcal{H}$}_{y_t}, \forall t = 1,2, ..., T\) . The conditional model is defined as:

\begin{equation}
\begin{split}
\label{eqn1}
P\left(\textbf{y} \mid \textbf{x}, \theta\right) = \sum_\textbf{h} P\left(\textbf{y} \mid \textbf{x}, \textbf{h}, \theta\right) P\left(\textbf{h} \mid \textbf{x}, \theta\right).
\end{split}
\end{equation}

Equation \eqref{eqn1} can be re-written using the graph structure in Fig. \ref{fig:CDCRF}c as:

\begin{equation}
\begin{split}
\label{eqn1-new}
P\left(\textbf{y} \mid \textbf{x}, \theta\right) = \sum_{\textbf{h}:\forall h_t \in \textit{$\mathcal{H}$}_{y_t} } P\left(\textbf{y} \mid \textbf{h}, \theta\right) P\left(\textbf{h} \mid \textbf{x}, \theta\right) \\ +
\sum_{\textbf{h}:\exists h_t \not\in \textit{$\mathcal{H}$}_{y_t} } P\left(\textbf{y} \mid \textbf{h}, \theta\right) P\left(\textbf{h} \mid \textbf{x}, \theta\right).
\end{split}
\end{equation}

Applying model constraints, we can write the following:
\vspace{2mm}
\begin{equation}
P({y_{t}}=\ell \mid {h_t}) = 
\begin{cases}
1, & h_t \in \mathcal{H}_{y_{t}=\ell} \\
0, & h_t \not\in \mathcal{H}_{y_{t}=\ell}.
\end{cases}
\label{eqn1-constraint}
\end{equation}

The model in equation \eqref{eqn1-new} can be simplified using equation \eqref{eqn1-constraint} as:

\begin{equation}
\label{eqn3}
P\left(\textbf{y} \mid \textbf{x}, \theta\right) = \sum_{\textbf{h}:\forall h_t \in \textit{$\mathcal{H}$}_{y_t} } P\left(\textbf{h} \mid \textbf{x}, \theta\right).
\end{equation}

$P\left(\textbf{h}\mid\textbf{x}, \theta\right)$ is described using Conditional Random Field formulation given by:

\begin{equation}
\label{eqn4}
P\left(\textbf{h} \mid \textbf{x}, \theta\right) = \frac{1}{\textbf{\textit{Z}}\left(\textbf{x},\theta\right)} \exp \left(\sum_k \theta_k . F_k\left(\textbf{h},\textbf{x}\right)\right),
\end{equation}

\noindent where index $k$ ranges over all parameters $\theta = \{\theta_k\}$ and \( \textbf{\textit{Z}}(\textbf{x},\theta) \) is the partition function defined as:

\begin{equation}
\label{eqn5}
\textbf{\textit{Z}}(\textbf{x},\theta) = \sum_\textbf{h} {\exp\left(\sum_k \theta_k . F_k(\textbf{h},\textbf{x})\right)}.
\end{equation}

\noindent The feature functions \(F_k \)'s are defined as: \[ F_k(\textbf{h},\textbf{x}) = \sum_{t=1}^T f_k(h_{t-1}, h_t, \textbf{x}, \textit{t}),\] 

Feature functions \(f_k(h_{t-1}, h_t, \textbf{x}, \textit{t}) \) can be either an \textit{observation} (also called \textit{state}) function \(s_k(h_t, \textbf{x}, \textit{t}) \) or a \textit{transition} function \(t_k(h_{t-1}, h_t, \textbf{x}, \textit{t}) \).

\subsection{Training Model Parameters} \label{Training}

Parameters of the model can be estimated by maximizing the conditional log-likelihood of the training data given by equation \eqref{eqn6}:

\begin{equation}
\label{eqn6}
\textit{\textbf{L}}(\theta) = \sum_{n=1}^N {\log P(\textbf{y}^{(n)} \mid \textbf{x}^{(n)}, \theta)} - \frac{{\parallel\theta\parallel}^2}{2\sigma^2},
\end{equation}

\noindent where \textit{N} is the total number of available labeled sequences. The second term in equation \eqref{eqn6} is the log of a Gaussian prior with variance \(\sigma^2\).

\subsection{Inference}

Given a new test sequence \textbf{x}, the inference task is given by:

\begin{equation}
\label{eqn7}
\hat{\textbf{y}} = \text{argmax}_{\textbf{y}} \quad P(\textbf{y} \mid \textbf{x}, \hat{\theta}),
\end{equation}

Using model constraints, equation \eqref{eqn7} can be re-written as:

\begin{equation}
\label{eqn8}
\hat{\textbf{y}} = \text{argmax}_{\textbf{y}} \quad \sum_{\textbf{h}:\forall h_t \in \textit{$\mathcal{H}$}_{y_t} } P(\textbf{h} \mid \textbf{x}, \hat{\theta}),
\end{equation}

We apply forward recursions of belief propagation for our inference as the problem of intention prediction must be solved online. In other words, at each time instant $t$, we compute the marginals $P(h_t \mid {x_{1:t}}, \theta)$ and sum them according to the disjoint sets of hidden states to obtain $P(y_t \mid {x_{1:t}}, \theta)$ = $\sum_{h_t \in \textit{$\mathcal{H}$}_{y_t} } P(h_t \mid {x_{1:t}}, \theta), t = 1,2, ...$. Then, we infer the label $y_t$ corresponding to the maximum probability. Forward-backward algorithm \cite{HMM} and Viterbi algorithm \cite{Viterbi} can also be applied for problems where online inference is not required. \\

\vskip 0.2in

\end{document}